\documentclass[lettersize,journal]{IEEEtran}
\usepackage{soul} 
\usepackage{color, xcolor} 
\usepackage{amsmath,amsfonts}
\usepackage{colortbl}
\definecolor{spink}{rgb}{0.99, 0.91, 0.95}
\definecolor{sblue}{rgb}{0.94, 0.97, 1.0}
\definecolor{syellow}{rgb}{1.0, 0.55, 0.0}
\usepackage{xspace}
\usepackage{array}
\usepackage{multicol}

\usepackage{stfloats}
\usepackage{url}
\usepackage{verbatim}
\usepackage{cite}
\usepackage{booktabs}
\usepackage{sidecap}
\usepackage{multirow}
\usepackage{floatrow}
\usepackage{wrapfig}
\usepackage{amssymb}
\usepackage{graphicx}
\usepackage{lipsum}
\usepackage{tikz}
\usepackage{pgfplots}

\usepackage{bbding} 
\usepackage{threeparttable}
\usepackage{tabularx} 
\usepackage{arydshln}
\usepackage{algorithm}
\usepackage{algorithmic}

\usepackage{mathrsfs}
\definecolor{darkgreen}{rgb}{0, 0.7, 0}
\definecolor{darkred}{rgb}{0.6, 0, 0}

\iftrue 
  \newcommand{\bassam}[1]{\noindent}
  \newcommand{\oscar}[1]{\noindent}
  \newcommand{\alex}[1]{\noindent}
  \newcommand{\sourabh}[1]{\noindent}
  \newcommand{\done}[1]{\noindent}
  \newcommand{\todo}[1]{\noindent}
\else
  \newcommand{\bassam}[1]{\textcolor{blue}{\bf [BH: #1]}}
  \newcommand{\oscar}[1]{\textcolor{orange}{\bf [OB: #1]}}
  \newcommand{\alex}[1]{\textcolor{purple}{\bf [AL: #1]}}
  \newcommand{\sourabh}[1]{\textcolor{brown}{\bf [SV: #1]}}
  \newcommand{\done}[1]{\textcolor{darkgreen}{\bf [Done: #1]}}
  \newcommand{\todo}[1]{\textcolor{red}{\bf [Todo: #1]}}
\fi
\usepackage[normalem]{ulem} 

\newcommand{\greenbf}[1]{\textcolor{darkgreen}{\bf #1}}

\newcommand{\gray}[1]{\textcolor{gray}{#1}}

\usetikzlibrary{patterns}
\pgfplotsset{compat=1.8}
\pgfmathdeclarefunction{fpumod}{2}{%
    \pgfmathfloatdivide{#1}{#2}%
    \pgfmathfloatint{\pgfmathresult}%
    \pgfmathfloatmultiply{\pgfmathresult}{#2}%
    \pgfmathfloatsubtract{#1}{\pgfmathresult}%
    \pgfmathfloatifapproxequalrel{\pgfmathresult}{#2}{\def\pgfmathresult{5}}{}%
}
\pgfplotsset{boxplot legend/.style={
    legend image code/.code={
        \draw[#1] (0cm,0cm) rectangle (0.6cm,0.3cm)
        (0.3cm,0cm) -- (0.3cm,-0.1cm) (0.1cm,-0.1cm) -- (0.5cm,-0.1cm)
        (0.3cm,0.3cm) -- (0.3cm,0.4cm) (0.1cm,0.4cm) -- (0.5cm,0.4cm);
    },
}}
 
\usepgfplotslibrary{statistics}
\usetikzlibrary{pgfplots.statistics}
\usepackage{subfigure}
\usepackage{lineno,hyperref}
\pgfplotsset{width=7cm,compat=1.8}
\pgfplotsset{%
  colormap={whitered}{color(0cm)=(white);
  color(1cm)=(blue!75!red)}
}
 \usepackage{pgfplotstable} 

\ifCLASSINFOpdf

\else
 
\fi

\usepackage{etoolbox}
\makeatletter
\patchcmd{\@makecaption}
  {\scshape}
  {}
  {}
  {}
\makeatother

\begin{document}
\title{Leveraging Anchor-based LiDAR 3D Object Detection via Point Assisted Sample Selection}


\author{Shitao Chen$^\ast$, Haolin Zhang$^\ast$ and Nanning Zheng{$^\dagger$}, \textit{Fellow, IEEE}

\thanks{$\ast$ Shitao Chen (E-mail: chenshitao@xjtu.edu.cn) and Haolin Zhang (E-mail: zhanghaolin@xjtu.edu.cn) contributed equally to this work. $\dagger$ Corresponding author: Nanning Zheng (E-mail: nnzheng@mail.xjtu.edu.cn).}
\thanks{Authors are with The National Key Laboratory of Human-Machine Hybrid Augmented Intelligence, National Engineering Research Center for Visual Information and Applications, and Institute of Artificial Intelligence and Robotics, Xi'an Jiaotong University, Xi'an, Shaanxi, China.}

}

\maketitle


\begin{abstract}
3D object detection based on LiDAR point cloud and prior anchor boxes is a critical technology for autonomous driving environment perception and understanding. Nevertheless, an overlooked practical issue in existing methods is the ambiguity in training sample allocation based on box Intersection over Union (IoU$_{box}$). This problem impedes further enhancements in the performance of anchor-based LiDAR 3D object detectors. To tackle this challenge, this paper introduces a new training sample selection method that utilizes point cloud distribution for anchor sample quality measurement, named Point Assisted Sample Selection (PASS). This method has undergone rigorous evaluation on two widely utilized datasets. Experimental results demonstrate that the application of PASS elevates the average precision of anchor-based LiDAR 3D object detectors to a novel state-of-the-art, thereby proving the effectiveness of the proposed approach. The codes will be made available at  \url{https://github.com/XJTU-Haolin/Point_Assisted_Sample_Selection}.       
\end{abstract}
\IEEEpeerreviewmaketitle

\section{Introduction}
\label{sec:intro}
\IEEEPARstart{B}{ackground}: In automated systems, such as autonomous driving and intelligent transportation, it is crucial to perceive and understand 3D surroundings, wherein 3D/bird's-eye view (BEV) object detection serves as a pivotal computer vision task. Owing to the accurate depth measurement capabilities of LiDAR sensors, 3D object detection using LiDAR-captured points has received widespread attention in recent years. According to learning objectives~\cite{review}, LiDAR-based 3D object detection can be categorized into anchor-based and anchor-free methods. Anchor-based methods~\cite{voxelnet,std,second,pointpillar,pillarnet,hvnet,cia-ssd,ago-net,ssn,ass-3dd,voxel-transformer,voxelrcnn,sassd,hvpr,pyrcnn,infofocus,voxeset,whatyousee} rely on predefined anchor boxes with fixed shapes, aiming to make predictions based on these anchor boxes and the input point clouds. In contrast, anchor-free methods~\cite{pixor,centerpoint,objectdgcnn,hotspot,pointrcnn,3dssd,pillar-OD,pvrcnnplus,octr,pvgnet} focus on directly predicting 3D objects without using predefined anchor boxes. Intuitively, the anchor-based models were presumed to be more easily learned than anchor-free models due to the prior knowledge of anchor boxes, including the uniform shapes within the same category and preset anchor positions covering the entire range of detection. However, with the development of the anchor-free methods~\cite{centerpoint,pillarnext,voxelnext}, their performance has gained momentum over the anchor-based methods, as evident in public 3D object detection benchmarks~\cite{waymo,nus}.  

\begin{figure}[t]
\begin{center}
\includegraphics[width = 1.0\columnwidth]{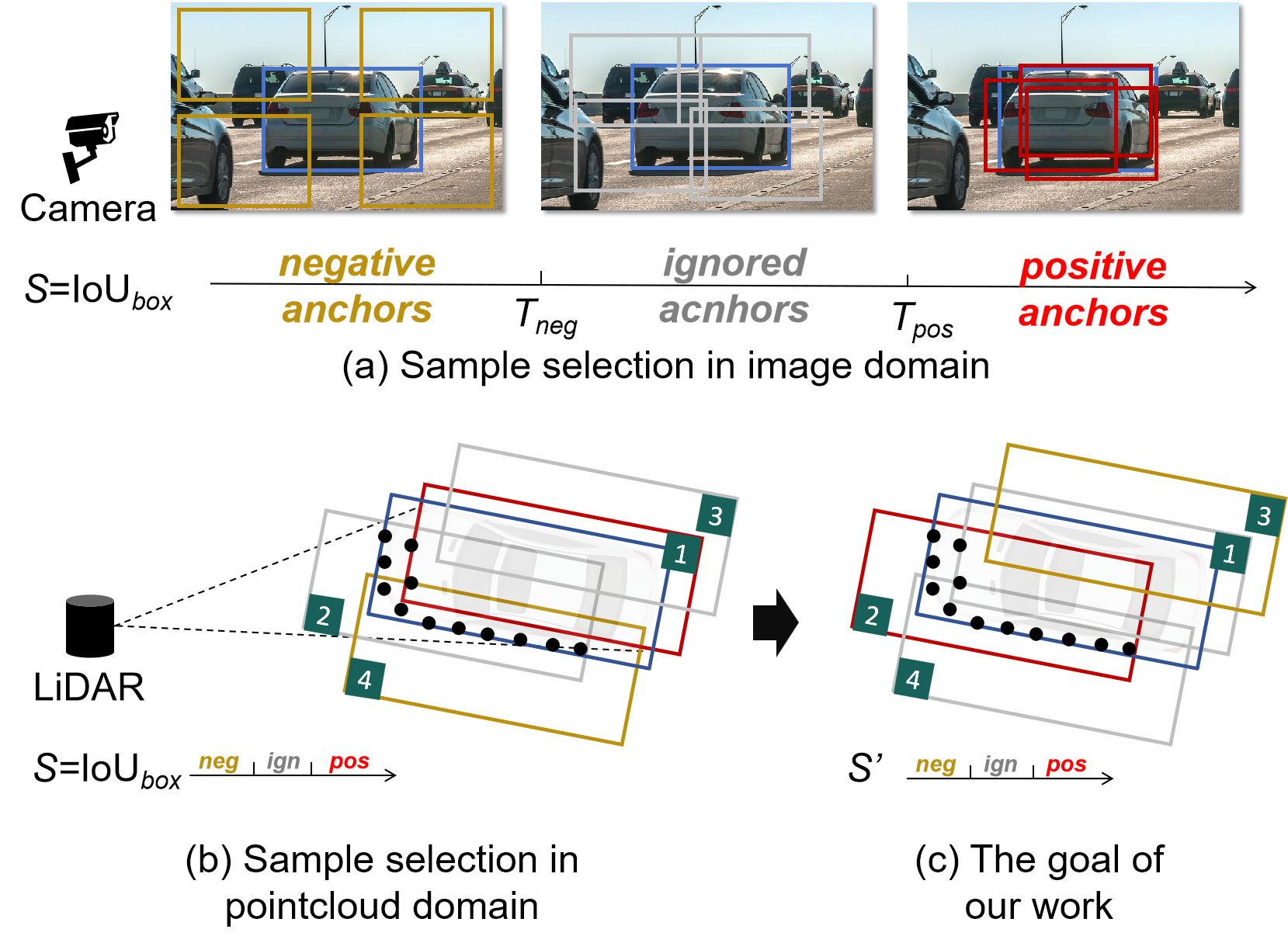}
\end{center}
\caption{Illustration of anchor sample selection in different modalities (\textcolor{blue}{blue boxes} indicate objects, best viewed in color). Anchor sample selection is required to assign samples into three subsets: \textcolor{red}{positive}, \textcolor{orange}{negative}, and \textcolor{gray}{ignored}. {(a) Sample selection in image domain}: the ambiguity of sample selection according to different measurement thresholds ($\mathcal{T}_{pos}^{}$ and $\mathcal{T}_{neg}^{}$) is small, as the sample selection metric $\mathcal{S}$ = IoU$_{box}$ primarily reflects the range of object pixel features contained by the anchor. { (b) Sample selection in pointcloud domain}: due to the sparsity nature of the LiDAR pointcloud, the same IoU$_{box}$-based sample selection scheme encounters greater ambiguity. For example, although anchor No.1 was selected as a positive sample with a high IoU$_{box}$, it lacked sufficient object point features compared to the ignored anchor No.2. Similarly, compared to the ignored anchor No.3, selecting anchor No.4, despite with more object point features, as a negative sample solely due to its lower IoU${box}$ is ambiguous. (c) To mitigate the ambiguity of sample selection in pointcloud domain, the goal is to design a novel form of the sample selection metric ($\mathcal{S^{\prime}}$) that better suits the learning of object point features. }
\label{fig:intro}
\end{figure}


\textbf{Motivation}: Due to the deployment-friendly nature, some anchor-based methods (e.g. PointPillar\cite{pointpillar}), accelerated by CUDA, have gained popularity in industrial applications. In recent years, researchers have proposed numerous methods that can enhance the accuracy of anchor-based detectors. Some methods employ multi-modalities fusion~\cite{pointpaint,SFD,virconv,transeq,logonet,vsc}, feature representations enhancement~\cite{PDV,FSC} or second-stage refinement network cascade~\cite{part2,lidarrcnn,pvrcnn,PDV}. These methods greatly elevate the performance of anchor-based LiDAR 3D object detectors. However, they often introduce additional data throughput or network parameters, thereby increasing the computational complexity of models and the costs of model deployment. Some other methods utilize general optimization techniques such as elaborating data augmentation~\cite{rs-aug}, employing knowledge distillation~\cite{itkd,unid} or designing novel loss functions~\cite{iouloss1,iouloss2,harmonicloss}. These methods do not fundamentally leverage the potential of anchor-based detectors themselves. To essentially refresh anchor-based LiDAR 3D object detectors, and narrow the performance gap between anchor-based and anchor-free detectors, it is necessary to analyze the limitations of existing anchor-based detectors.

\textbf{Analysis and Inference}: What restricts the basic performance of anchor-based LiDAR 3D object detection methods? To answer this question, we delve deeper into reviewing and analyzing the existing LiDAR-based detectors' learning objectives, by which whether a detector belongs to anchor-based or anchor-free is fundamentally determined. 

The major difference in learning objectives between anchor-based and anchor-free methods lies in the selection of positive and negative training samples. As shown in Figure~\ref{fig:intro} (a) and (b), the widely-used anchor-based LiDAR 3D object detection methods~\cite{pointpillar,second,pvrcnn,FSC,PDV} all adopt a training sample assignment strategy based on Intersection over Union of boxes (IoU$_{box}$), similar to 2D anchor-based object detection~\cite{f-rcnn,ssd} in the image domain. IoU$_{box}$ is an efficient means to define the degree of overlap between two bounding boxes, thereby depicting the spatial similarity between an anchor sample and an object. Given the dense pixel distribution in the image space, IoU$_{box}$ also predominantly signifies the number of semantic features encompassed by the anchor, thereby establishing a clear learning objective and facilitating sufficient feature learning. However, in the pointcloud domain, the LiDAR-captured points typically exhibit sparsity due to factors such as the field of view angle, number of scan lines, and occlusion. The sparsity of points results in incomplete object representations. Consequently, IoU$_{box}$ fails to measure the completeness of the semantic features contained within the anchor sample, which introduces ambiguity into the learning objective when relying only on IoU$_{box}$ to assign training samples in anchor-based LiDAR 3D object detectors. 
 
 Without the involvement of anchors, the selection of training samples in anchor-free LiDAR 3D object detection methods does not exhibit significant ambiguity. Depending on the specific model framework, there are four typical categories of the sample selection strategy in anchor-free methods: grid-based~\cite{pixor,pillar-OD,centerpoint,pillarnext,voxelnext}, point-based~\cite{pointrcnn,3dssd}, range-based~\cite{lasernet,rangedet}, and set-to-set assignment~\cite{voxeset,endendtr3d,octr}. Although these methods differ in processing details, they offer little ambiguity in learning objectives. This is achieved by constraining positive samples to be located within or near the object center, converting sparse pointclouds into dense pseudo-images, or establishing one-to-one sample matching relationships. This indicates that a clear selection metric is important, and the core is to ensure that the selection metrics can unequivocally assess the sample quality.

From the above analysis, two inferences are drawn as follows: (1) The relatively high density of image pixels guarantees that IoU$_{box}$ can effectively define the quality of an anchor sample with small ambiguity, as it reflects the spatial and semantic similarity between the anchor box and the object box. However, this is not always the case with sparse LiDAR pointclouds. (2) Anchor-free LiDAR 3D object detection methods execute sample selection based on the objects themselves or pseudo-dense representations, leading to clearer sample selection measurement and less ambiguous feature learning compared to anchor-based methods. 

\textbf{Problem Definition}: The existing IoU$_{box}$-based sample assignment strategy in anchor-based LiDAR 3D object detection introduces ambiguity in the learning objective, and hence limits the basic performance of anchor-based methods from the optimization. The primary challenge of releasing the ambiguity hinges on how to define a clearer selection metric for dividing positive and negative samples, which is the goal of this work as shown in figure~\ref{fig:intro}(c). 

\textbf{Proposed Solutions}: To address the ambiguity of sample selection and leverage the performance of anchor-based LiDAR 3D object detectors, a series of solutions were proposed in this work. Firstly, a specific sample quality measurement,  IoU$_{point}$ (Intersection over Union of points in two cubes), is proposed by further analyzing the limitations of IoU$_{box}$-based sample selection in anchor-based LiDAR 3D object detection. Statistic analysis and case studies are conducted to illustrate the defects of a single IoU$_{box}$ measurement and the significance of incorporating IoU$_{point}$ measurement. Secondly, a novel training sample selection approach, termed \textit{\textbf{P}oint \textbf{A}ssisted \textbf{S}ample \textbf{S}election} (PASS), is proposed. PASS integrates IoU$_{box}$ and IoU$_{point}$ to provide a clearer assessment of anchor samples, facilitating unambiguous feature learning. Finally, extensive experiments are conducted on two widely-used datasets (the KITTI dataset~\cite{kitti} and Waymo Open Dataset~\cite{waymo}), demonstrating that the application of PASS promotes the performance of anchor-based 3D object detectors. These results underscore the effectiveness of the proposed solutions. The contributions of this work are summarized into three parts:

(1) The learning ambiguity of IoU$_{box}$-based sample selection is pointed out for the first time in the research field of anchor-based LiDAR 3D object detection. To better capture the semantic features contained within anchor samples in LiDAR pointcloud domain, a novel sample quality measurement, IoU$_{point}$, is proposed. Statistical analysis and case studies are conducted for illustration and analysis. 

(2) A tailed training sample assignment method, named PASS, is proposed. This method offers a clearer selection metric between positive and negative anchor samples during model optimization, effectively reducing learning ambiguity.  

(3) Comparative experiments on several large-scale datasets validate the effectiveness of our proposed method. From a practical perspective, PASS can be plug-and-play to train any anchor-based LiDAR 3D object detector, without introducing additional model parameters and inference time-cost.

The paper continues as Section \ref{related} that provides an overview of related research. Section \ref{review} reviews and analyzes the existing IoU$_{box}$-based sample selection, followed by an illustration of the proposed IoU$_{point}$ measurement and its significance. Section \ref{proposed} elaborates on the proposed training sample assignment method PASS. In Section~\ref{ex}, the effectiveness of the proposed method is evaluated through qualitative and quantitative analysis of 3D object detection experiments. Section \ref{sec5} outlines the conclusion.


\section{Related Work}
\label{related}

LiDAR sensors provide accurate depth estimation and fine-grained 3D object surfaces, surpassing the capabilities of 2D cameras. LiDAR 3D object detection methods, with the input of LiDAR pointcloud, normally predict the 3D bounding boxes and classes of surrounding objects. These methods are categorized into anchor-based and anchor-free approaches according to their learning objectives, specifically the paradigm of training sample determination.

Anchor-based LiDAR 3D object detection methods require the pre-definition of prior 3D anchor boxes within the 3D space. Some methods adopt a one-stage framework, such as VoxelNet~\cite{voxelnet} and SECOND~\cite{second}, which involves converting LiDAR pointcloud into 3D voxels and applying 3D convolutions for feature extraction and object detection. PointPillars~\cite{pointpillar} simplifies the 3D voxels into 2D pillars for efficiency. PillarNet~\cite{pillarnet} is proposed to improve these one-stage anchor-based detectors with a novel design of a pillar-based 3D detection network. Some other methods introduce a two-stage network to build a from-coarse-to-fine framework~\cite{pvrcnn,part2,pvrcnnplus,FSC,graph-rcnn,pg-rcnn}. The first stage predicts the 3D proposals based on prior anchors, and the subsequent stage aims to refine the proposals to output the final detection results. Despite variations in network or algorithm design among anchor-based methods, their learning objectives remain the same: classifying the object based on features around prior anchors and regressing the bounding boxes from the positive anchor samples. These methods commonly rely on Intersection over Union of boxes (IoU$_{box}$) to assign anchors to the positive, ignored, or negative set. Such an IoU$_{box}$-based measurement makes little ambiguity in image 2D object detection~\cite{fasterrcnn}, since IoU$_{box}$ roughly measures the spatial and semantic similarity at the same time in the context of dense pixels. Some studies~\cite{noisyanchor,atss,autoass} have even extended the IoU$_{box}$-based measurement to further improve the effectiveness of model training. However, such a selection method meets a great ambiguity in the LiDAR pointcloud domain. Although IoU$_{box}$ can approximately measure the spatial similarity between the two boxes, the completeness of object features contained by the anchor cannot be accurately represented due to the uneven distribution and sparsity of LiDAR points. This ambiguity potentially leads to the fact that some anchors with even more semantic features may be discarded because of the only IoU$_{box}$-based measurement, while low-quality anchors may be learned as positive samples, affecting the accuracy and training convergence of models. 

Anchor-free LiDAR 3D object detection methods exhibit less ambiguity in learning objectives, compared to anchor-based methods. PIXOR~\cite{pixor} uses grid cells in ground truths as the measurement of positive training samples. Such an inside-based assignment strategy is also adopted in \cite{pillar-OD}. CenterPoint~\cite{centerpoint} builds a Gaussian kernel at each center of objects to indicate a positive label. This center-based assignment strategy is applied to the latest anchor-free detectors~\cite{pillarnext,voxelnext}. Point-based assignment strategy~\cite{pointrcnn,3dssd} is also popularly adopted in anchor-free detectors. The points are first segmented into foreground and background, where the foreground points inside or near the ground truths are selected as positive samples for training. These assignment methods all share a common advantage: through object points or object centers, a clear positive and negative division boundary is defined, which reduces the ambiguity of network learning. In addition, transformer-style detectors~\cite{voxeset,endendtr3d,octr} introduce a set-to-set assignment approach, which performs a mapping from each positive sample to an object by Hungarian matching, keeping a one-to-one unambiguous manner. Range-based assignment strategy~\cite{lasernet,rangedet} selects the positive samples inside the objects, which does not bring a big ambiguity because it works on the relatively dense range images generated by LiDAR pointcloud. 

Based on the above review and analysis, the learning objective of existing anchor-based LiDAR 3D object detection methods is not fully reasonable because of the ambiguity of IoU$_{box}$-based anchor sample selection in LiDAR pointcloud. Resolving this ambiguity may help improve the effectiveness of feature learning and the detection performance of anchor-based detectors.

\section{Analysis of Sample Selection in Anchor-based LiDAR 3D Object Detection}
\label{review}

This section begins by providing the preliminary of existing $\text{IoU}_{box}$-based sample selection scheme. Subsequently, a hypothesis of the ambiguity of $\text{IoU}_{box}$-based sample selection in LiDAR pointcloud domain is presented and explained. Following a new $\text{IoU}_{point}$-based selection measurement is proposed and illustrated. Finally, the static analysis and case studies are carried out to further demonstrate the ambiguity of $\text{IoU}_{box}$-based sample selection and the importance of inducting an $\text{IoU}_{point}$-based assessment.

\subsection{Preliminary of Existing $\text{IoU}_{box}$-based Sample Selection}
\label{review-pre}

3D object detection aims to predict the cuboids (3D bounding boxes) that enclose the ground truth objects. Each cuboid can be represented as $
\left[ x^g,y^g,z^g,l^g,w^g,h^g,\theta ^g \right]$, where $\{x^g,y^g,z^g\}$ indicate the 3D center coordinates, $\{l^g,w^g,h^g\}$ denote the length, width and height of the cuboid, and $\theta ^g$ means the orientation of the object.
Anchor-based LiDAR 3D object detection relies on pre-defined cubids with fixed shapes, represented as $
\left[ x^a,y^a,z^a,l^a,w^a,h^a,\theta ^a \right]$. They are placed in the 3D point cloud space as the prior anchor boxes. These anchors are assigned to the positive, negative, or ignored set via the training sample selection procedure. The learning objective entails recognizing the class of objects and predicting their center, shape, and orientation from the positive anchors. The negative anchors are classified as background. The ignored anchors remain untrained, acting as fuzzy samples that delineate the positive and negative boundaries. 


Existing state-of-the-art anchor-based LiDAR 3D object detection methods (e.g. \cite{second,pointpillar,part2,pvrcnn,FSC,pg-rcnn}) rely on IoU$_{box}$-based training sample selection. For a pair of ground truth $\text{box}_{gt}$ and anchor $\text{box}_{anchor}$, a selection metric $\mathcal{S}^{(g,a)}$ is directly determined by IoU$_{box}$ as Eq.~\ref{eq:ioub}.

\begin{equation}
{\mathcal{S}^{(g,a)}} = \text{IoU}_{box}=\frac{\mathcal{I}_{volume}\left( \text{box}_{gt},\text{box}_{anchor} \right)}{\mathcal{U}_{volume}\left( \text{box}_{gt},\text{box}_{anchor}  \right) }  \in [0,1]
\label{eq:ioub}
\end{equation}
Where $\mathcal{I}_{volume}()$ represents the volume intersection of cuboids, and $\mathcal{U}_{volume}()$ represents the volume union of cuboids.

For each object category, a positive threshold $\mathcal{T}_{pos}^{}$ and a negative threshold $\mathcal{T}_{neg}^{}$ are pre-defined. Each anchor is assigned according to $\mathcal{S}^{(g,a)}$ by Eq.~\ref{eq:assign}. 


\begin{equation}
\begin{cases}
	{\text{box}_{anchor}}{\rightarrow}\text{positive} \,\,\,\,   if\,\,{\mathcal{S}^{(g,a)}}>{\mathcal{T}_{pos}}\\
	\text{box}_{anchor}{\rightarrow}\text{negtive}\,\,             \,\,else\,\,if\,\,{\mathcal{S}^{(g,a)}}<{\mathcal{T}_{neg}}\\
	{\text{box}_{anchor}}{\rightarrow}\text{ignored}\,\,\,\,            otherwise\,\, \\
\end{cases}
\label{eq:assign}
\end{equation}

where \{$\mathcal{T}_{pos}^{}$,$\mathcal{T}_{neg}^{}$\} is manually set. In existing anchor-based methods' settings, \{$\mathcal{T}_{pos}^{}$,$\mathcal{T}_{neg}^{}$\} are generally set to \{0.6,0.45\} for the car class, and \{0.5,0.35\} for the pedestrian class and the cyclist class.

\begin{figure}[t!]
\begin{center}
\includegraphics[width = 1.0\columnwidth]{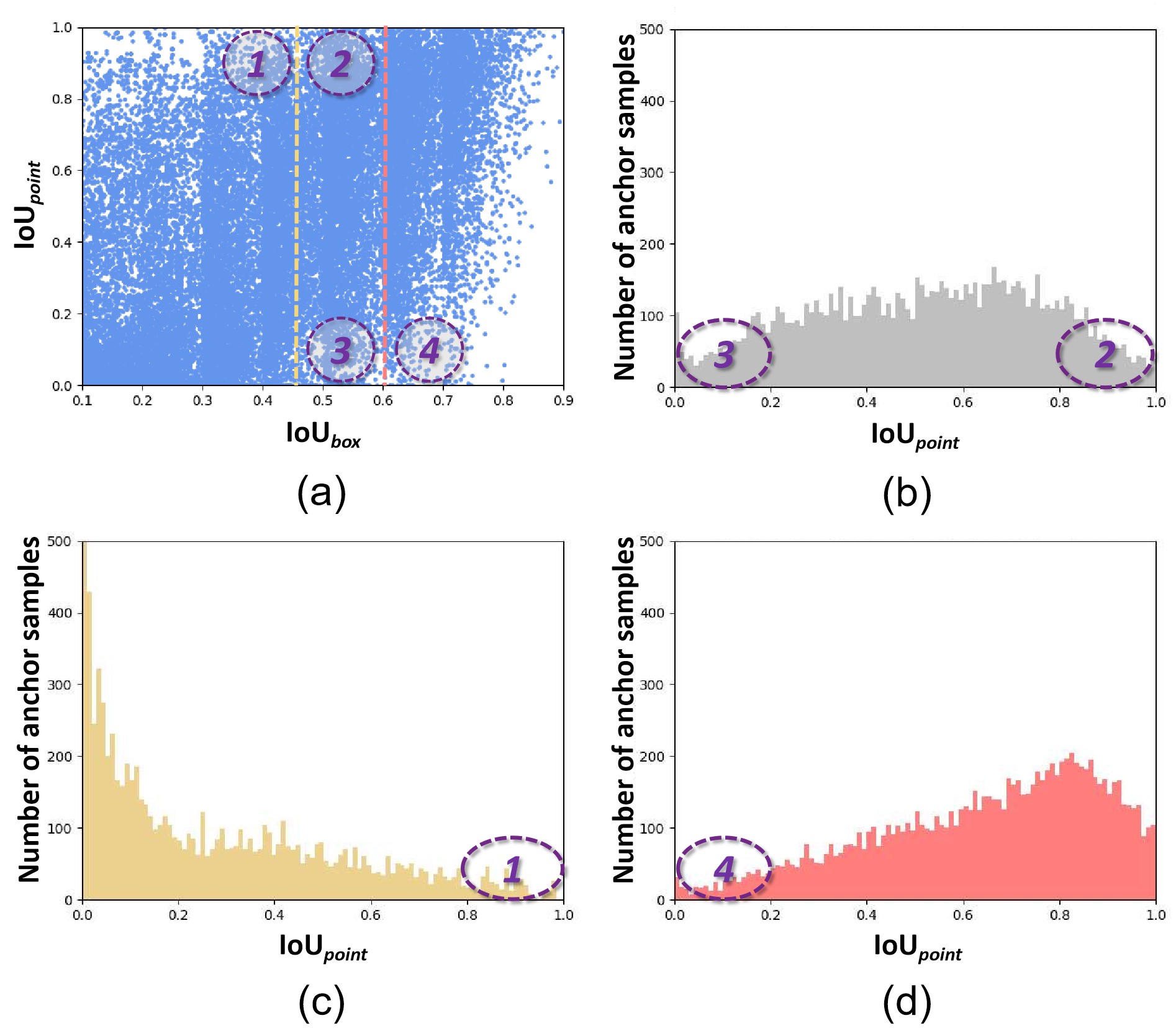}
\end{center}
    \caption{Statistics of training anchor samples distribution over the KITTI dataset. (a) The scatter
diagram of the relation between IoU$_{box}$ and IoU$_{point}$. (b) The histogram diagram of IoU$_{point}$ of those ignored anchor samples determined by IoU$_{box}$-based sample selection scheme. (c) The histogram of IoU$_{point}$ of those negative anchor samples determined by IoU$_{box}$-based sample selection scheme. (d) The histogram of IoU$_{point}$ of those positive anchor samples determined by IoU$_{box}$-based sample selection scheme.}

\label{fig:dis}
\end{figure}

\subsection{Hypothesis and Proposed $\text{IoU}_{point}$-based Measurement}

In image object detection, pixels serve as the primitive semantic units for feature learning. Since image pixels are relatively dense, $\text{IoU}_{box}$ can be broadly interpreted as a measure of both spatial and semantic similarity between the anchor and the ground truth object, leading to minimal ambiguity in training sample selection and subsequent feature learning, as depicted in Fig.\ref{fig:intro}(a). Comparatively, as shown in Fig.~\ref{fig:intro}(b), LiDAR points, as the main source of input data and original semantic features for LiDAR 3D object detection, are often sparse and partially absent. In pointcloud domain, $\text{IoU}_{box}$ can still reflect the spatial approximation of the anchor box to the ground truth box. However, it fails to depict the semantic approximation of the anchor to the ground truth due to the sparse distribution of scanned points associated with the object. Consequently, a hypothesis emerges regarding such inconsistency: $\text{IoU}_{box}$ potentially brings about ambiguity when it acts as the only measurement for training sample assignment in LiDAR anchor-based 3D object detection. An intuitive way to verify and solve this ambiguity is taking into account the similarity of pointcloud as an additional measurement alongside $\text{IoU}_{box}$. Referring to the form and characteristics of $\text{IoU}_{box}$, a new measurement $\text{IoU}_{pointcloud}$, as seen in Eq.\ref{eq:ioup}, that depicts the overlap of two sets of pointcloud is proposed. 

\begin{equation}
\text{IoU}_{point}=\frac{\mathcal{I}_{{\#}points}\left( \text{box}_{gt},\text{box}_{anchor} \right)}{\mathcal{U}_{{\#}points}\left( \text{box}_{gt},\text{box}_{anchor} \right)} \in [0,1]
\label{eq:ioup}
\end{equation}
Where $\mathcal{I}_{{\#}points}()$ represents the number of points enclosed in the intersection of cuboids, and $\mathcal{U}_{{\#}points}()$ represents the number of points included in the union of cuboids.

\begin{figure}[t!]
\begin{center}
\includegraphics[width = 1.0\columnwidth]{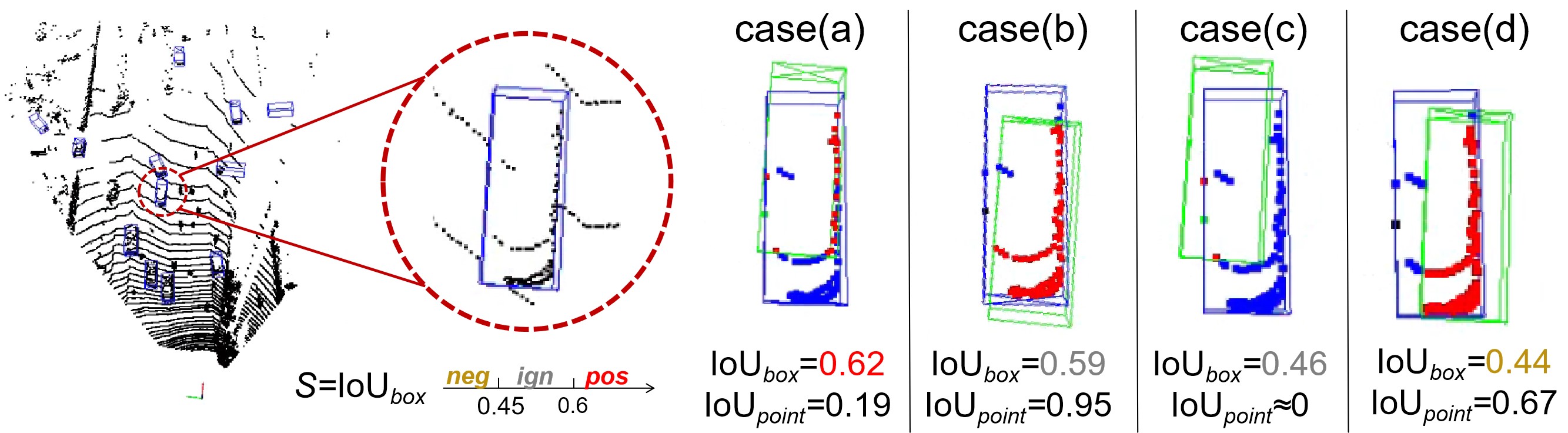}
\end{center}
    \caption{Visualized case studies of ambiguity associated with IoU$_{box}$-based sample selection (best viewed in color). The \textcolor{green}{green boxes} and \textcolor{blue}{blue boxes} represent the anchor samples and ground truths respectively. Points in the intersection of the anchor sample and ground truth are indicated as \textcolor{red}{red points}. Points that belong to ground truth but do not belong to the intersection are indicated as \textcolor{blue}{blue points}. Points that belong to the anchor but do not belong to the intersection are represented as \textcolor{green}{green points}.}
\label{fig:case}
\end{figure}

\subsection{Statistic analysis and case studies}

To further analyze the effect of sample selection in anchor-based LiDAR 3D object detection, the statistics on approximately one thousand training anchor samples for cars from the KITTI dataset~\cite{kitti} are drawn as in figure~\ref{fig:dis}. For better illustration and explanation, some typical cases are presented visually in figure~\ref{fig:case}. In figure~\ref{fig:dis}(a), each scatter represents an anchor, and the horizontal and vertical coordinates represent IoU$_{box}$ and IoU$_{point}$ between the anchor and its corresponding ground truth, respectively. We plotted the red dotted line to represent the threshold $\mathcal{T}_{pos}^{}$ for positive samples and the yellow dotted line to represent the threshold $\mathcal{T}_{neg}^{}$ for negative samples. The existing IoU$_{box}$-based sample selection method separates these anchors directly by comparing the $\mathcal{S}$ = IoU$_{box}$ to \{$\mathcal{T}_{pos}^{}$,$\mathcal{T}_{neg}^{}$\}. However, by observing the highlighted regions as numbered in purple circles, the hypothesized ambiguity happens. For instance, some anchors, even with high IoU$_{point}$ indicated in the area of the No.1 circle, are assigned as the negative samples because of the relatively low IoU$_{box}$. Conversely, anchors with low IoU$_{point}$ in the area of the No.3 circle are assigned to the ignored set only because of the slightly higher IoU$_{box}$. Such ambiguity is further elucidated through comparisons between case(c) and case(d) in figure~\ref{fig:case}. Moreover, some anchors with high IoU$_{box}$ but low IoU$_{point}$ are assigned to the positive set as in the area of the No.4 circle. However, the anchors with similar IoU$_{box}$ but very high IoU$_{point}$ are assigned to the ignored set as in the area of the No.2 circle. Such ambiguity is clarified by comparing case(a) and case(b) in figure~\ref{fig:case}. It can be seen that solely relying on IoU$_{box}$ fails to provide a comprehensive assessment of the quality of an anchor sample, especially regrading to the completeness of the object features. Consequently, IoU$_{box}$-based sample selection method creates inherent learning ambiguity for anchor-based LiDAR 3D object detectors. The histogram diagrams of IoU$_{point}$ of ignored, negative, and positive anchor samples determined by IoU$_{box}$-based sample selection scheme are shown in figure~\ref{fig:dis}(b),  (c), and (d), respectively. From these distributions, it is evident that the proportion of the anchors with selection ambiguity mentioned above is non-negligible. Addressing this ambiguity is crucial, and introducing IoU$_{point}$ into the sample selection measurement is the main idea of our solution, which will be detailed in the next section.

\section{Proposed Solution: Point Assisted Sample Selection}
\label{proposed}

Based on the aforementioned analysis, this section introduces a novel method for sample selection in anchor-based LiDAR 3D object detection, named Point Assisted Sample Selection (PASS). This method incorporates the IoU$_{point}$ metric alongside the IoU$_{box}$ metric used in prior works to eliminate sample assignment ambiguity.

It is empirically unreasonable to use the average of IoU$_{box}$ and IoU$_{point}$ as the selection measurement. This is because, for some objects with very sparse points, the value of IoU$_{point}$ is significantly smaller than that of IoU$_{box}$, potentially resulting in an insufficient number of positive training samples. Consequently, first a upper boundary $\mathcal{B}_{upper}$ and a lower boundary $\mathcal{B}_{lower}$ are defined based on the selection thresholds \{$\mathcal{T}_{pos}^{}$,$\mathcal{T}_{neg}^{}$\} and a hyperparameter $\mathcal{K}$, as shown in Eq.~\ref{eq:T}. The value of $\mathcal{K}$ governs the dynamic margin of boundaries. \{$\mathcal{B}_{upper}$,$\mathcal{B}_{lower}$\} determines a specified interval, covering the positive and negative selection thresholds, as the adjustment range to which PASS is applied.

\begin{equation}
\left\{ \begin{aligned}
	\mathcal{B}_{upper}&=\mathcal{T}_{pos}^{}+\small{\frac{1}{\mathcal{K}}}\left( \mathcal{T}_{pos}^{}-\mathcal{T}_{neg} \right)\\
	\mathcal{B}_{lower}&=\mathcal{T}_{neg}^{}-\small{\frac{1}{\mathcal{K}}}\left( \mathcal{T}_{pos}^{}-\mathcal{T}_{neg} \right)\\
\end{aligned} \right. 
\label{eq:T}
\end{equation}

The previous IoU$_{box}$-based measurement $\mathcal{S}$=IoU$_{box}$ retains its significance in PASS, as it provides an approximate assessment of the spatial similarity between the anchor and the ground truth. IoU$_{point}$, serving as a measure of point feature similarity, can be considered as an assisted metric to adjust $\mathcal{S}$. In PASS, the new selection measurement, denoted as  $\mathcal{S}^{\prime}$, is defined as shown in  Eq.\ref{eq:assign}. In practical applications, it is observed that many background points originating from the ground plane are emcompassed within the 3D bounding boxes of objects or anchors, causing unreasonable values of IoU$_{point}$. To mitigate this issue, before the calculation of IoU$_{point}$, removing points from the ground plane is recommended. For this purpose, the RANSAC-based plane segmentation algorithm from the open-source 3D data processing library Open3D\footnote{Open3D \url{https://github.com/isl-org/Open3D}} is employed in this work.

\begin{equation}
\begin{aligned}
\mathcal{S}^{\prime} &=\alpha\mathcal{S}+ \beta(\text{IoU}_{point} \mathcal{B}_{upper} 
+\left( 1-\text{IoU}_{point} \right)  \mathcal{B}_{lower} )\\
&=\alpha\text{IoU}_{box}+ \beta(\text{IoU}_{point}\mathcal{B}_{upper}+\left( 1-\text{IoU}_{point} \right)  \mathcal{B}_{lower} ) 
\end{aligned} 
\label{eq:pass}
\end{equation}
where $\alpha$ and $\beta$ are hyperparameters that weighting the IoU$_{box}$-based measurement and IoU$_{point}$-assisted measurement, and $\alpha$+$\beta$=1. IoU$_{box}$-based sample selection measurement in current anchor-based LiDAR 3D object detectors can be considered as a special form of PASS by setting $\alpha$=1 and $\beta$=0. In this work, we simply average the weights of two metrics by set $\alpha$=$\beta$=$\frac{1}{2}$. It not only eliminates the redundant hyperparameters, but leads to boundary constraints as follows:

For $ {\forall} \mathcal{S}\leqslant \mathcal{T}_{neg}^{}$, following Eq.~\ref{eq:T} and Eq.~\ref{eq:pass}, $\mathcal{S}^{\prime}$ can be bounded as Eq.~\ref{eq:left}.
\begin{equation}
\begin{aligned}
\mathcal{S}^{\prime} &\leqslant \small{\frac{1}{2}}\left( \mathcal{T}_{neg}^{}+\mathcal{B}_{upper} \right) \,\,    {\text{since}} \ \text{IoU}_{point}\leqslant 1\\
	&=\small{\frac{1}{2}}\left( \mathcal{T}_{neg}^{}+\mathcal{T}_{pos}^{} \right) +\small{\frac{1}{2\mathcal{K}}}\left( \mathcal{T}_{pos}^{}-\mathcal{T}_{neg} \right)    \\
	&\leqslant \small{\frac{1}{2}}\left( \mathcal{T}_{neg}^{}+\mathcal{T}_{pos}^{} \right) +\small{\frac{1}{2}}\left( \mathcal{T}_{pos}^{}-\mathcal{T}_{neg} \right) \,\, {\text{since}} \ \mathcal{K} \geqslant 1\\
	&=\,\,\mathcal{T}_{pos}^{} \,\,\,\,\\
\end{aligned}
\label{eq:left}
\end{equation}

The above inequality proves that the samples that were originally classified into the negative set by IoU$_{box}$-based measurement will not cross into the positive set by PASS, that is, they will remain in the negative set or cross over to the ignored set.

Similarly, for $ {\forall} \mathcal{S}\geqslant \mathcal{T}_{pos}^{}$, following Eq.~\ref{eq:T} and Eq.~\ref{eq:pass}, $\mathcal{S}^{\prime}$ can be bounded as Eq.~\ref{eq:right}.
\begin{equation}
\begin{aligned}
\mathcal{S}^{\prime} &\geqslant \small{\frac{1}{2}}\left( \mathcal{T}_{pos}^{}+\mathcal{B}_{lower} \right) \,\,  {\text{since}} \ \text{IoU}_{point}\geqslant 0\\
	&=\small{\frac{1}{2}}\left( \mathcal{T}_{pos}^{}+\mathcal{T}_{neg}^{} \right) -\small{\frac{1}{2\mathcal{K}}}\left( \mathcal{T}_{pos}^{}-\mathcal{T}_{neg} \right) \,\, \\
	&\geqslant \small{\frac{1}{2}}\left( \mathcal{T}_{pos}^{}+\mathcal{T}_{neg}^{} \right) -\small{\frac{1}{2}}\left( \mathcal{T}_{pos}^{}-\mathcal{T}_{neg} \right) \,\,{\text{since}} \ \mathcal{K} \geqslant 1\\
	&=\,\,\mathcal{T}_{neg}^{}\\
	\,\,\\
\end{aligned}
\label{eq:right}
\end{equation}

The above inequality proves that the samples that were originally classified into the positive set by IoU$_{box}$-based measurement will not cross into the negative set by PASS. They will be kept in the positive set or assigned to the ignored set.

For $\forall \mathcal{S}$ that  $\mathcal{T}_{neg}^{}\leqslant \mathcal{S}\leqslant \mathcal{T}_{pos}^{}$, following Eq.~\ref{eq:T} and Eq.~\ref{eq:pass}, $\mathcal{S}^{\prime}$ can be bounded as Eq.~\ref{eq:middle1} and Eq.~\ref{eq:middle2}.

\begin{equation}
\begin{aligned}
 \mathcal{S}^{\prime}&\leqslant \small{\frac{1}{2}}\left( \mathcal{T}_{pos}^{}+\mathcal{B}_{upper} \right) \,\, {\text{since}} \ \text{IoU}_{point}\leqslant 1\\
	&=\small{\frac{1}{2}}\left( \mathcal{T}_{pos}^{}+\mathcal{T}_{pos}^{}+\small{\frac{1}{K}}\left( \mathcal{T}_{pos}^{}-\mathcal{T}_{neg} \right) \right)\\
	&=\mathcal{T}_{pos}^{}+\small{\frac{1}{2\mathcal{K}}}\left( \mathcal{T}_{pos}^{}-\mathcal{T}_{neg} \right) \,\,\\
	&\leqslant \mathcal{T}_{pos}^{}+\small{\frac{1}{\mathcal{K}}}\left( \mathcal{T}_{pos}^{}-\mathcal{T}_{neg} \right) \,\, {\text{since}} \ \mathcal{K}\geqslant 1\\
	&=\,\,\mathcal{B}_{upper}\\
\end{aligned}
\label{eq:middle1}
\end{equation}

\begin{equation}
\begin{aligned}
 \mathcal{S}^{\prime}&\geqslant \small{\frac{1}{2}}\left( \mathcal{T}_{neg}^{}+\mathcal{B}_{lower} \right) \,\, {\text{since}} \ \text{IoU}_{point}\geqslant 0\\
	&=\small{\frac{1}{2}}\left( \mathcal{T}_{neg}^{}+\mathcal{T}_{neg}^{}-\small{\frac{1}{\mathcal{K}}}\left( \mathcal{T}_{pos}^{}-\mathcal{T}_{neg} \right) \right)\\
	&=\mathcal{T}_{neg}^{}-\small{\frac{1}{2\mathcal{K}}}\left( \mathcal{T}_{pos}^{}-\mathcal{T}_{neg} \right) \,\,\\
	&\geqslant \mathcal{T}_{neg}^{}-\small{\frac{1}{\mathcal{K}}}\left( \mathcal{T}_{pos}^{}-\mathcal{T}_{neg} \right) \,\, {\text{since}} \ \mathcal{K}\geqslant 1\\
	&=\,\,\mathcal{B}_{lower}\\
\end{aligned}
\label{eq:middle2}
\end{equation}

The above inequalities prove that for the samples that were originally classified into the ignored set, \{$\mathcal{B}_{upper}$,$\mathcal{B}_{lower}$\} strictly constrain the value of PASS measurement $\mathcal{S}^{\prime}$. 

These constraints ensure that the purpose of PASS is to adjust the assignment of ambiguous samples caused by the IoU$_{box}$-based metric while avoiding the introduction of new ambiguity. Algorithm~\ref{alg:pass} shows the overall process of how to select positive and negative anchor samples via PASS.

\begin{algorithm}[t!]
\small
\caption{Point Assisted Sample Selection (PASS)} 
\label{alg:pass} 
\begin{algorithmic}[1]
\REQUIRE ~~\\
$\mathcal{G}$ is a set of ground-truth boxes \\
$\mathcal{A}$ is a set of all anchor boxes \\
$\mathcal{K}$ is a hyperparameter of PASS\\
$\mathcal{T}_{pos}$ is a threshold for selecting positive sample\\
$\mathcal{T}_{neg}$ is a threshold for selecting negative sample\\
\ENSURE ~~\\
$\mathcal{P}$ is a set of positive samples \\
$\mathcal{N}$ is a set of negative samples \\
\vspace{2mm}
\STATE compute $\mathcal{B}_{lower}$,$\mathcal{B}_{upper}$ based on $\{\mathcal{T}_{pos},\mathcal{T}_{neg},\mathcal{K}\}$; (See Eq.~\ref{eq:T}) 
\FOR{each ground-truth box $g \in \mathcal{G}$}
\FOR{each anchor box $a \in \mathcal{A}$}
\STATE compute $\mathcal{S}^{(g,a)} = \text{IoU}^{{(g,a)}}_{box}$; (See Eq.~\ref{eq:ioub})\\
\IF{ $ \mathcal{B}_{lower} \leqslant \mathcal{S}^{(g,a)} \leqslant \mathcal{B}_{upper}$ }
    \STATE compute  $\text{IoU}^{(g,a)}_{point}$; (See Eq.~\ref{eq:ioup})
    \STATE compute $\mathcal{S^{\prime}}^{(g,a)}$ based on \\
    $\{ \mathcal{S}^{(g,a)}, \text{IoU}^{(g,a)}_{point}, \mathcal{B}_{lower}, \mathcal{B}_{upper} \}$ ; (See Eq.~\ref{eq:pass})
    \STATE update $\mathcal{S}^{(g,a)} = \mathcal{S^{\prime}}^{(g,a)}$ ; 
    \STATE record $\mathcal{S}^{(g,a)}$ $\rightarrow$ $\mathcal{S}^{(\mathcal{G},\mathcal{A})}$;
\ELSE{  
\STATE record $\mathcal{S}^{(g,a)}$ $\rightarrow$ $\mathcal{S}^{(\mathcal{G},\mathcal{A})}$;}

\ENDIF
\ENDFOR
\ENDFOR

\STATE collect $\mathcal{P}$ and $\mathcal{N}$ based on $\{\mathcal{T}_{pos},\mathcal{T}_{neg},\mathcal{S}^{(\mathcal{G},\mathcal{A})}\}$; (See Sec.~\ref{review-pre})
\RETURN $\mathcal{P}, \mathcal{N}$;
\end{algorithmic}
\end{algorithm}



\section{Experiments and Analysis}
\label{ex}
In this section, extensive experiments on multiple datasets are conducted to evaluate the effectiveness of the proposed method. Firstly, the experimental datasets and implementation details are introduced. Secondly,  comparative experiments are performed to validate the feasibility of the proposed method. Furthermore, the proposed method is applied to an advanced anchor-based detector, and compared to state-of-the-art detectors on benchmarks. Subsequently, ablation studies are conducted to evaluate the effects of different configurations. Finally, qualitative analysis and discussion are provided.

\begin{table*}[]
\caption{Verification of applying PASS to anchor-based LiDAR 3D object detectors on the KITTI \textit{val} set. 
}
\small
\center
\renewcommand{\arraystretch}{0.85}  
\setlength\tabcolsep{6.0pt}  
\begin{tabular}{| c | c|c || c | c | c || c | c | c || c | c | c |}
\hline
\multirow{2}{*}{Method}   & \multirow{2}{*}{\#Stages} & Avg. & \multicolumn{3}{|c||}{Car (IoU=0.7)} & \multicolumn{3}{|c||}{Pedestrian (IoU=0.5)} & \multicolumn{3}{|c|}{Cyclist (IoU=0.5)} \\ \cline{3-12}
  &    & Mod. & Easy   & Mod.   & Hard   & Easy      & Mod.     & Hard     & Easy     & Mod.    & Hard      \\ \hline \hline






SECOND~\cite{second}   & One	 &69.84 & 89.76	& 87.05	  &\textbf{85.09}	& 57.81	 & 52.48	   	& 48.33	  & 81.76	              			& 69.99	  & \textbf{66.09}	   
\\

\textbf{+ PASS}  			& One	 	 &\textbf{71.22}	& \textbf{89.82}	  & \textbf{87.25}	 & 85.04	       & \textbf{60.00}	 & \textbf{55.74}	 & \textbf{50.69}					& \textbf{82.22}	 & \textbf{70.68}	 & 65.74	  		 	\\  
\hdashline
$\Delta$ 		 			& -	  &\greenbf{+1.38} &  \greenbf{+0.06}  & \greenbf{+0.20}  & \gray{-0.05}          		& \greenbf{+2.19}  & \greenbf{+3.26}  & \greenbf{+2.36}					& \greenbf{+0.46}  & \greenbf{+0.69} & \gray{-0.35}		\\ \hline\hline

PointPillars~\cite{pointpillar}   & One   &70.57  & \textbf{89.70}	& \textbf{87.40}	  & 83.57  & \textbf{61.40}	  & 56.49	    		& \textbf{51.61}	   & 82.83  & 67.81	  & 63.00	  
\\

\textbf{+ PASS} 			& One		 &\textbf{70.87}	& 89.43	  &87.38	& \textbf{85.03}        & 59.76	 & \textbf{56.53}	 & 51.17					& \textbf{83.29}	 & \textbf{68.69}	 & \textbf{63.76}				\\  
\hdashline
$\Delta$ 		 			& -	  &\greenbf{+0.30} &  \gray{-0.27}  & \gray{-0.02}  & \greenbf{+1.46}          		& \gray{-0.64}  & \greenbf{+0.04}  & \gray{-0.44}					& \greenbf{+0.46}  & \greenbf{+0.88} & \greenbf{+0.76}		\\ \hline\hline 

PillarNet~\cite{pillarnet}     & One	  	 &68.53 		& 89.73  & 86.89  & 84.27     		& 56.79  & 52.40  & 48.79            			& \textbf{84.05}  & \textbf{66.29} & \textbf{62.30}     		\\
\textbf{+ PASS}   		 & One		 &\textbf{69.64}		& \textbf{89.85}  & \textbf{87.43}  & \textbf{85.19}          	& \textbf{60.06}  & \textbf{55.29} & \textbf{51.81}				& 83.65  & 66.21  & 61.92  			\\ \hdashline
$\Delta$ 		 	& -				 &\greenbf{+1.11}	 &  \greenbf{+0.12}  & \greenbf{+0.54}   & \greenbf{+0.92}          		&\greenbf{+3.27}  & \greenbf{+2.89}  & \greenbf{+3.02}					&  \gray{-0.40}  &  \gray{-0.08} &  \gray{-0.38}		
\\

\hline \hline

PV-RCNN~\cite{pvrcnn}  & Two	  &72.91  & 89.89	& 87.55   		& 86.37  & \textbf{67.21}  & 58.67     		& 54.50  & 87.64   & \textbf{72.50}  & 69.07   
\\
\textbf{+ PASS} 		 & Two	    &\textbf{73.41}		& \textbf{90.11}  & \textbf{88.02}  & \textbf{87.45}         	& 66.53  & \textbf{60.34}  & \textbf{55.21}				& \textbf{91.05}  & 71.88  & \textbf{69.33}  		\\  
\hdashline
$\Delta$ 		 		& -		 &\greenbf{+0.50} &  \greenbf{+0.22}  & \greenbf{+0.47}  & \greenbf{+1.08}          		& \gray{-0.68}  & \greenbf{+1.67}  & \greenbf{+0.69}					& \greenbf{+2.41}  & \gray{-0.62} & \greenbf{+0.26}		\\ \hline\hline  

Focals-Conv~\cite{FSC}     & Two	  	 &73.56  		& \textbf{90.18}  & \textbf{88.11}  & \textbf{87.52}     		& 68.17  & 60.81  & 55.18            			& 85.08  & 71.75  & 67.08     		\\
\textbf{+ PASS}  		 & Two		 &\textbf{74.35} 		& 89.98  & 87.41  & 86.87          	& \textbf{68.98}  & \textbf{61.60} & \textbf{56.39}				& \textbf{87.64}  & \textbf{74.05}  & \textbf{70.73}  			\\ \hdashline
$\Delta$ 		 	& -				 &\greenbf{+0.79}	 &  \gray{-0.20}  & \gray{-0.70}  & \gray{-0.65}          		& \greenbf{+0.81}  & \greenbf{+0.79}  & \greenbf{+1.21}					& \greenbf{+1.36}  & \greenbf{+2.30} & \greenbf{+3.65}		\\ \hline\hline

Graph-Vo~\cite{graph-rcnn}     & Two	  	 &75.66 		& 90.25  & 88.76  & 87.83    		& 70.53  & 64.06  & 58.21            			& 86.96  & \textbf{74.17}  & \textbf{72.72}    		\\
\textbf{+ PASS}  		 & Two		&\textbf{76.05}  &\textbf{90.47}		&\textbf{88.99}  &\textbf{88.17}  &\textbf{71.46}          	&\textbf{65.45}  &\textbf{59.51} 	&\textbf{86.99} &73.71  &71.18			\\ \hdashline
$\Delta$ 		 	& -		& \greenbf{+0.39}		 &\greenbf{+0.22} &\greenbf{+0.23} &\greenbf{+0.34}  &\greenbf{+0.93}          		& \greenbf{+1.39} &\greenbf{+1.30} & \greenbf{+0.03}					& \gray{-0.46} & \gray{-0.46}\\ \hline\hline 

PV-RCNN++~\cite{pvrcnnplus}     & Two	  	 &74.06  		& \textbf{90.24}  & \textbf{88.14}  & \textbf{87.61}    		& 65.53  & 59.28  & 55.03            			& \textbf{94.81}  & 74.76  & \textbf{70.79}    		\\
\textbf{+ PASS}  		 & Two		&\textbf{75.00}  &90.11  &88.10		  &87.60       &\textbf{66.86}  	&\textbf{62.00}   &\textbf{56.84}	&93.76 &\textbf{74.91}  &70.65			\\ \hdashline
$\Delta$ 		 	& -		& \greenbf{+0.94} 	& \gray{-0.13}		 &\gray{-0.04} &\gray{-0.01} &\greenbf{+1.33}  &\greenbf{+2.72}          		& \greenbf{+1.81} &\gray{-1.05} & \greenbf{+0.15}					& \gray{-0.14} \\

\hline
\end{tabular}

\begin{tablenotes}
        \footnotesize
        \item  $\#$Stages: number of stages.  Avg.: Average. Mod.: Moderate. Best results are highlighted in \textbf{bold}.
\end{tablenotes}

\vspace{1mm}

\label{table:kt_val_3d}
\end{table*}


\begin{table*}[]
\caption{Verification of applying PASS to anchor-based LiDAR 3D object detectors (use 20\% training data) on the Waymo Open Dataset \textit{val} set. 
}
\small
\center
\renewcommand{\arraystretch}{0.85}  
\setlength\tabcolsep{6.0pt}  
\begin{tabular}{| c | c|c || c | c  || c | c  || c | c |}
\hline
\multirow{2}{*}{Method}  & \multirow{2}{*}{$\#$Stages} & mAPH& \multicolumn{2}{|c||}{VEH (AP/APH)} & \multicolumn{2}{|c||}{PED (AP/APH)} & \multicolumn{2}{|c|}{CYC (AP/APH)}  \\ \cline{3-9}
  &    & L1/L2 & L1   & L2   & L1   & L2     & L1    & L2       \\ \hline \hline





SECOND~\cite{second}   & One   &60.0/54.3  & \textbf{71.0}/\textbf{70.3}	& \textbf{62.6}/\textbf{62.0}	  & 65.2/54.2  &57.2/47.5	  & 57.1/55.6	    & 55.0/53.5   	   
\\

\textbf{+ PASS} 			& One	 	 &\textbf{61.1}/\textbf{55.3} & 70.3/69.7	  &61.9/61.4	& \textbf{66.9}/\textbf{56.0}        & \textbf{58.8}/\textbf{49.1}	 & \textbf{59.1}/\textbf{57.6}	 & \textbf{56.9}/\textbf{55.4}								\\  
\hdashline
$\Delta$ 		& -	  &  \greenbf{+1.1}/\greenbf{+1.0}  			&\gray{-0.7}/\gray{-0.6} &  \gray{-0.7}/\gray{-0.6}  & \greenbf{+1.7}/\greenbf{+1.8}  
&\greenbf{+1.6}/\greenbf{+1.6}    & \greenbf{+2.0}/\greenbf{+2.0}  & \greenbf{+1.9}/\greenbf{+1.9} 
\\ \hline\hline

PointPillars~\cite{pointpillar}   & One   &56.0/50.7  & 70.4/69.8	& 62.2/61.6	  & \textbf{66.2}/46.3  &58.2/40.6	  & 55.3/51.8	    & 53.2/49.8	   	   
\\

\textbf{+ PASS} 			& One	 	 &\textbf{58.0}/\textbf{52.6} & \textbf{70.6}/\textbf{69.9}	  &\textbf{62.3}/\textbf{61.7}	& \textbf{66.2}/\textbf{47.7}        & \textbf{58.4}/\textbf{42.0}	 & \textbf{58.9}/\textbf{56.3}	 & \textbf{56.7}/\textbf{54.1}								\\  
\hdashline
$\Delta$ 		 			& -	 &\greenbf{+2.0}/\greenbf{+1.9} &  \greenbf{+0.2}/\greenbf{+0.1}  & \greenbf{+0.1}/\greenbf{+0.1}  
&NA/\greenbf{+1.4}    & \greenbf{+0.2}/\greenbf{+1.4}  & \greenbf{+3.6}/\greenbf{+4.5}  & \greenbf{+3.5}/\greenbf{+4.3}							\\ \hline\hline 

PV-RCNN~\cite{pvrcnn}  & Two	   &66.7/60.9  & 75.4/\textbf{74.7}	& 67.4/\textbf{66.8}	  & 72.0/61.2  &63.7/54.0	  & 65.9/64.3	    & 63.4/61.8     
\\

\textbf{+ PASS} 			& Two	& \textbf{67.3}/\textbf{61.5}	 &\textbf{75.5}/\textbf{74.7} & \textbf{67.5}/\textbf{66.8}	  &\textbf{72.8}/\textbf{61.7}	&\textbf{64.5}/\textbf{54.5}        & \textbf{67.4}/\textbf{65.6}	 & \textbf{64.9}/\textbf{63.2}\\  
\hdashline
$\Delta$ 		 	& -		 & \greenbf{+0.6}/\greenbf{+0.6} 	 &\greenbf{+0.1}/NA  &  \greenbf{+0.1}/NA   & \greenbf{+0.8}/\greenbf{+0.5}  & \greenbf{+0.8}/\greenbf{+0.5}    & \greenbf{+1.5}/\greenbf{+1.3}  & \greenbf{+1.5}/\greenbf{+1.4}  \\ \hline\hline  

PV-RCNN++~\cite{pvrcnn}  & Two	   &68.8/62.7  & \textbf{77.0}/\textbf{76.5}	& \textbf{69.3}/\textbf{68.8}	  &69.9/61.3 &60.8/53.1	  & 70.0/68.7	    & 67.5/66.3    
\\

\textbf{+ PASS} 			& Two	& \textbf{69.1}/\textbf{63.3} &76.8/76.3 &69.2/68.6	  &\textbf{70.7}/\textbf{61.9}	&\textbf{62.7}/\textbf{54.7}        &\textbf{70.6}/\textbf{69.1}	 & \textbf{68.1}/\textbf{66.7}\\  
\hdashline
$\Delta$ 		 	& -		& \greenbf{+0.3}/\greenbf{+0.4} & \gray{-0.2}/\gray{-0.2}	 &\gray{-0.1}/\gray{-0.2}  &  \greenbf{+0.8}/\greenbf{+0.6}   & \greenbf{+1.9}/\greenbf{+1.6} & \greenbf{+0.6}/\greenbf{+0.4}    & \greenbf{+0.6}/\greenbf{+0.4}   
				
\\ \hline 


\end{tabular}

\begin{tablenotes}
        \footnotesize
        \item  $\#$Stages: number of stages. VEH: Vehicle. PED: Pedestrian. CYC: Cyclist. NA: not applicable. Best results are highlighted in \textbf{bold}.
\end{tablenotes}

\vspace{1mm}

\label{table:waymo_val_3d}
\end{table*}

\begin{table*}[]
\caption{Comparison with state-of-the-art LiDAR 3D object detectors on KITTI \textit{test} benchmark.
}
\small
\center
\renewcommand{\arraystretch}{0.85}  
\setlength\tabcolsep{3.0pt}  
\begin{tabular}{| c | c| c|c || c | c | c ||c | c | c || c | c | c ||   c |}
\hline
\multirow{2}{*}{Method}  & \multirow{2}{*}{$\#$Stages} & \multirow{2}{*}{Anchor}  & \multicolumn{1}{|c||}{Avg.} & \multicolumn{3}{|c||}{Car} & \multicolumn{3}{|c||}{Pedestrian} & \multicolumn{3}{|c||}{Cyclist} & \multirow{2}{*}{Ref.} \\ \cline{4-13}
  &  &  & Mod. & Easy   & Mod.   & Hard   & Easy      & Mod.   & Hard    & Easy      & Mod.   & Hard  &    \\ \hline \hline


3DSSD~\cite{3dssd} & One &No  &62.65  & 88.36   		&79.57  & 74.55  & 54.64 	&44.27 	&40.23  &82.48 &64.10 &56.90      	 & CVPR'20
\\
HotSpotNet~\cite{hotspot} & One &No   &63.21 &87.60  		&78.31    &73.34  & 53.10		&45.37  &41.47    &82.59 &65.95 &59.00        	 & ECCV'20
\\


IA-SSD~\cite{iassd} & One &No  &62.53  & 88.87   		&80.32  & 75.10  & 47.90 		& 41.03  & 37.98   &82.36 &66.25 &59.70       	 & CVPR'22
\\

PointRCNN~\cite{pointrcnn}  & Two &No    & 57.94 	& 86.96   		&75.64  &70.70  & 47.98   		& 39.37 & 36.01     &74.96 &58.82 & 52.53    			   & CVPR'19
\\ 



\hline\hline
PointPillar~\cite{pointpillar} & One &Yes  & 58.29  &82.58   		&74.31  & 68.99  &51.45	& 41.92  &38.89   &77.10 &58.65  &51.92       	 & CVPR'19
\\

SVGA-Net~\cite{svga-net} & One &Yes  &61.05   & 	87.33  		&80.47   &75.91  & 48.48		&40.39  &37.92   &78.58 &62.28 &54.88        	 & AAAI'22
\\
PillarNet~\cite{pillarnet} & One &Yes  &63.41  &89.65  		&81.06   &76.67  &46.71		&40.85  &38.54   &83.01 &68.33 &61.07        	 & ECCV'22
\\

TANet~\cite{tanet} & Two &Yes & 59.91 & 84.39   		&75.94  & 68.82  &\textbf{53.72}		& \textbf{44.34}  &\textbf{40.49}   &75.70 &59.44 &52.53       	 & AAAI'20
\\

Part-$A^2$~\cite{part2} & Two &Yes &61.70   & 87.81   		&78.49  & 73.51  &53.10		&43.35  & 40.06   &79.17 &63.25 &56.93        	 & TPAMI'20
\\
PV-RCNN~\cite{pvrcnn} & Two &Yes & 62.81  & 90.25  		&81.43  & 76.82  &52.17		& 43.29 & 40.29   &78.60 &63.71 &57.65       	 & CVPR'20
\\ 

PDV~\cite{PDV} & Two &Yes & 63.41 & \textbf{90.43}   		&81.86  & \textbf{77.36}  & 47.80		&40.56  &38.46   &\textbf{83.04} &67.81 &60.46        	 & CVPR'22
\\

PG-RCNN~\cite{pg-rcnn} & Two &Yes &63.66  & 89.38  		&\textbf{82.13}   &77.33  & 47.99		&41.04  &38.71   &82.77 &67.82 &\textbf{61.25}        	 & ICCV'23
\\



\hline 
PV-RCNN++~\cite{pvrcnnplus}  & Two &Yes & 63.02  & {87.72}   		&{81.29}  & 76.78  & 47.50   		&40.31 & 38.15     &80.34 &67.46 &60.48        			   & IJCV'23
\\
\textbf{PASS-PV-RCNN++} 		& Two &Yes  & \textbf{63.89}	 	& 87.65   		&81.28  & {76.79}  & {47.66}   		&{41.95} &{38.90}     &{80.43} &\textbf{68.45} &{60.93}		& - 	\\

\hdashline
$\Delta$ 		 	& -			 & - & \greenbf{+0.87}  & \gray{-0.07}          		& \gray{-0.01}  & \greenbf{+0.01}  & \greenbf{+0.16}	& \greenbf{+1.64} & \greenbf{+0.75}  &\greenbf{+0.09}  &\greenbf{+0.99}  & \greenbf{+0.45} 				& -	\\ 

\hline



\hline
\end{tabular}
\begin{tablenotes}
        \footnotesize
        \item  $\#$Stages: number of stages. Avg.: Average. Mod.: Moderate. Ref.:Reference. The best results among anchor-based methods are highlighted in \textbf{bold}.
        \item The listed results are cited from KITTI test server \url{https://www.cvlibs.net/datasets/kitti/eval_3dobject.php}~\cite{kitti}
\end{tablenotes}

\vspace{1mm}

\label{table:kt_sota_3class}
\end{table*}

\begin{table*}[]
\caption{Comparison with state-of-the-art LiDAR 3D object detectors on Waymo Open Dataset \textit{val} set.}
\small
\center
\renewcommand{\arraystretch}{0.85}  
\setlength\tabcolsep{2.0pt}  
\begin{tabular}{| c | c |c |c || c | c  || c | c  || c | c ||c |}
\hline
\multirow{2}{*}{Method}  & \multirow{2}{*}{$\#$Stages}  & \multirow{2}{*}{Anchor} & mAPH & \multicolumn{2}{|c||}{VEH (AP/APH)} & \multicolumn{2}{|c||}{PED (AP/APH)} & \multicolumn{2}{|c||}{CYC (AP/APH)} & \multirow{2}{*}{Ref.} \\ \cline{4-10}
     & &  & L1/L2 & L1   & L2   & L1   & L2     & L1    & L2   &     \\ \hline \hline





CenterPoint~\cite{centerpoint}   & One  &No   & 73.5/67.6 & 76.6/76.0	& 68.9/68.4	  &79.0/73.4  &71.0/65.8	  & 72.1/71.0	    & 69.5/68.5	  &CVPR'21 	   
\\ 

Graph-Ce~\cite{graph-rcnn}   & Two  &No  & 77.8/71.6  & 80.6/80.1	&72.3/71.9  & 82.9/77.3  &75.0/69.7	  & 77.2/76.0	    & 74.4/73.3		  &ECCV'22 	   
\\ 

PV-RCNN++$^\dagger$~\cite{pvrcnnplus}   & Two  &No  & 75.9/69.5 & 79.3/78.8	&70.6/70.2  &81.3/76.3  &73.2/68.0	  & 73.7/72.7	    & 71.2/70.2	  &IJCV'23 	   
\\ 

GD-MAE~\cite{gdmae}   & Two  &No  & 77.6/71.6 & 80.2/79.8	&72.4/72.0  &83.1/76.7  &75.5/69.4	  & 77.2/76.2	    & 74.4/73.4	  &CVPR'23 	   
\\ 


\hline\hline
SECOND~\cite{second}   & One &Yes  & 63.1/57.2 & 72.3/71.7	& 63.9/63.3  & 68.7/58.2  &60.7/51.3	  & 60.6/59.3	    & 58.3/57.1   &Sesnors'18	   
\\
PointPillars~\cite{pointpillar}   & One  &Yes  & 63.5/57.8 & 72.1/71.5	& 63.6/63.1	  & 70.6/56.7 &62.8/50.3	  & 64.4/62.3	    & 61.9/59.9	  &CVPR'19 	   
\\

PV-RCNN~\cite{pvrcnn}   & Two &Yes & 69.6/63.3  & 77.5/76.9	&69.0/68.4  & 75.0/65.6  &66.0/57.6	  & 67.8/66.4	    & 65.4/64.0   &CVPR'20	   
\\
Part-$A^2$~\cite{part2}   & Two &Yes & 70.3/63.8   & 77.1/76.5	& 68.5/68.0  & 75.2/66.9  &66.2/58.6	  & 68.6/67.4	    & 66.1/64.9   &TPAMI'20	   
\\

LiDAR-RCNN~\cite{lidarrcnn}   & Two &Yes & 67.0/61.3  & 76.0/75.5	& 68.3/67.9  & 71.2/58.7  &63.1/51.7	  & 68.6/66.9    & 66.1/64.4   &CVPR'21	   
\\

PDV~\cite{PDV}   & Two &Yes & 70.0/64.2  & 76.9/76.3	& 69.3/68.8  &74.2/66.0  &65.9/58.3	  & 68.7/67.6    & 66.5/65.4   &CVPR'22	   
\\ \hline
PV-RCNN++~\cite{pvrcnnplus}   & Two &Yes & 71.0/64.9  & 78.8/78.2	&70.3/69.7  &\textbf{76.7}/\textbf{67.2} &\textbf{68.5}/\textbf{59.7}	  & 69.0/67.6	    & 66.5/65.2  &IJCV'23	   
\\


\textbf{PASS-PV-RCNN++}			& Two	&Yes & \textbf{72.0}/\textbf{65.7}	 &\textbf{79.3}/\textbf{78.8} & \textbf{70.5}/\textbf{70.0}	  &76.2/66.9	& 67.2/58.8        &\textbf{71.8}/\textbf{70.7}	 & \textbf{69.4}/\textbf{68.3}	 & -								\\  
\hdashline
$\Delta$ 		 	& -			 & -  & \greenbf{+1.0}/\greenbf{+0.8} & \greenbf{+0.5}/\greenbf{+0.6}  & \greenbf{+0.2}/\greenbf{+0.3}         		& \gray{-0.7}/\gray{-0.3}  & \gray{-1.3}/\gray{-0.9}  & \greenbf{+2.8}/\greenbf{+3.1}	& \greenbf{+2.9}/\greenbf{+3.1}			& -	\\
\hline 

\end{tabular}

\begin{tablenotes}
        \footnotesize
        \item PV-RCNN++: anchor-based version of \cite{pvrcnnplus}. PV-RCNN++$^\dagger$: anchor-free version of \cite{pvrcnnplus}. $\#$Stages: number of stages. VEH: Vehicle. PED: Pedestrian. CYC: Cyclist. Ref.: Reference. The best results among anchor-based methods are highlighted in \textbf{bold}.
\end{tablenotes}

\vspace{1mm}

\label{table:waymo_sota}
\end{table*}

\begin{table*}[]
\caption{
Ablation study on the effect of hyperparameter $\mathcal{K}$ of PASS on the KITTI \textit{val} set. }
\small
\center
\renewcommand{\arraystretch}{1.0}  
\setlength\tabcolsep{0.8pt}  
\begin{tabular}{| c | c| c || c | c | c || c | c | c || c | c | c |}
\hline
\multirow{2}{*}{Method} &\multirow{2}{*}{$\mathcal{K}$}  & mAP  & \multicolumn{3}{|c||}{Car (IoU=0.7/0.5)} & \multicolumn{3}{|c||}{Pedestrian (IoU=0.5/0.25)} & \multicolumn{3}{|c|}{Cyclist (IoU=0.5/0.25)} \\ \cline{3-12}
& & Avg. & Easy   & Mod.   & Hard   & Easy      & Mod.     & Hard     & Easy     & Mod.    & Hard    \\ \hline \hline

Baseline    & -		& 64.44   		& 86.50/96.57  & 77.56/94.44  & 74.70/93.68     		& 52.59/73.14  & 47.01/69.26  & 42.87/65.38            			& 80.00/87.85  & 61.52/70.55  & 57.23/66.63      \\ \hdashline

 {+ PASS} 		& 2		& 65.43    		& 87.28/\textbf{97.36}  & 77.99/94.55  & 75.05/93.72       		& 53.47/\textbf{76.04}  & 47.80/\textbf{71.63}  &  44.13/\textbf{68.01}	        			& 81.87/88.97  & 62.74/71.65  & 58.60/67.49\\

 {+ PASS} 		& 3		&   64.95		& 87.62/95.75  & 78.23/94.52  & 75.07/93.47       		& 53.98/74.06  & 49.24/70.93  &  45.43/67.76		        			& 77.70/88.21  & 60.64/71.88  & 56.61/68.25\\

+ PASS 		& 4		&   65.55 		& 86.68/96.73  & 77.76/94.47  & 74.97/93.45        		& 54.89/73.93  & 49.19/70.45  & 44.81/66.44		        			& 80.05/\textbf{90.22}  & 62.94/73.51  & 58.63/68.90			\\ 

+ PASS 		& 5		&   \textbf{67.16}		& \textbf{87.74}/96.99  & \textbf{78.53}/94.49  & \textbf{75.60}/93.62        		& \textbf{57.72}/73.65  & \textbf{51.28}/70.76  & \textbf{46.53}/67.21		        			& \textbf{83.56}/89.26  & 63.96/70.95  & 59.53/66.91			\\

+ PASS 		& 6		&   65.88		& 87.33/95.67  & 78.34/\textbf{94.58}  & 75.39/\textbf{93.74}       		&  54.82/74.17  & 48.94/70.26  & 44.64/66.74		        			& 81.87/89.63  & 62.73/73.17  & 58.88/68.93			\\
+ PASS 		& 7		&   66.76		& 87.29/95.78  & 78.31/94.42  & 75.45/93.58       		&  56.06/73.79  & 50.09/69.46  & 45.55/65.82		        			& 82.30/89.10  & \textbf{65.06}/\textbf{74.18}  & \textbf{60.69}/\textbf{70.07}			\\
+ PASS 		& 8		&   64.65		& 87.03/95.54  & 77.82/94.36  & 75.06/93.41       		&  54.11/73.20  & 48.55/69.04  & 44.14/65.04		        			& 78.17/88.39  & 60.65/70.77  & 56.36/66.58			\\\hdashline

$\Delta_{avg}$  		& -		&   \greenbf{+1.33}		& \greenbf{+0.78}/ \gray{  -0.31}  & \greenbf{+0.58}/\greenbf{+0.04}   &  \greenbf{+0.53}/ \gray{  -0.11}      		& \greenbf{+2.42}/\greenbf{+0.98}   & \greenbf{+2.28}/\greenbf{+1.10}   &  \greenbf{+2.16}/\greenbf{+1.34}		        			& \greenbf{+0.79}/\greenbf{+1.26}  & \greenbf{+1.15}/ \greenbf{+1.75}  & \greenbf{+1.24}/\greenbf{+1.53}			\\

$\Delta_{max}$   		& -		&   \greenbf{+2.72}		& \greenbf{+1.24}/\greenbf{+0.79}  & \greenbf{+0.97}/\greenbf{+0.14}   &  \greenbf{+0.90}/\greenbf{+0.06}      		& \greenbf{+5.13}/\greenbf{+2.90}   & \greenbf{+4.27}/\greenbf{+2.37}   &  \greenbf{+3.66}/\greenbf{+2.63}		        			& \greenbf{+3.56}/\greenbf{+2.37}  & \greenbf{+3.54}/ \greenbf{+3.63}  & \greenbf{+3.46}/\greenbf{+3.44}			\\

 \hline 

\end{tabular}
\begin{tablenotes}
        \footnotesize
        \item Avg.: Average. Mod.: Moderate. $\Delta_{avg}$/$\Delta_{avg}$: the average/maximum mAP difference. The best results are highlighted in \textbf{bold}.
\end{tablenotes}

\vspace{1mm}

\label{table:kt_abb}
\end{table*}

\begin{table*}[tb]
\caption{
Ablation study on the anchor-based detector w/wo PASS($\mathcal{K}=5$)/Second-Stage on the KITTI \textit{val} set.}
\small
\center
\renewcommand{\arraystretch}{0.5}  
\setlength\tabcolsep{5.0pt}  
\begin{tabular}{| c| c||c|c | c||c | c | c|| c | c | c |}
\hline
 \multirow{2}{*}{PASS} &\multirow{2}{*}{2nd-S}  & \multicolumn{3}{|c||}{Car (IoU=0.7/0.5)} & \multicolumn{3}{|c||}{Pedestrian (IoU=0.5/0.25)} & \multicolumn{3}{|c|}{Cyclist (IoU=0.5/0.25)} \\ \cline{3-11}
 &  & Easy   & Mod.   & Hard   & Easy      & Mod.     & Hard  & Easy     & Mod.    & Hard    \\ \hline \hline
    \XSolidBrush   & \XSolidBrush 		 		&95.2/99.6 &90.6/95.9 &88.0/95.2 		&61.0/74.4 &55.2/71.7 &50.9/67.5            			&88.5/90.9 &69.7/72.0 &65.3/68.0      \\
    

   \CheckmarkBold  & \XSolidBrush 				& 95.8/99.4  &  90.9/96.0  & 88.4/95.3    		&61.7/76.1  & 56.6/72.6  &51.1/67.9            			& 86.5/88.7  &68.8/73.9  &63.4/69.7	\\

   \XSolidBrush		& \CheckmarkBold   	 	& 96.3/99.6  & 91.7/97.7  & 89.1/95.4    		& 70.6/82.9  & 64.1/79.6  & 59.2/\textbf{75.9}            			& \textbf{92.3}/\textbf{95.5}  & \textbf{74.4}/78.8  & \textbf{71.5}/74.5		\\
  \CheckmarkBold  &\CheckmarkBold		 		&\textbf{96.5}/\textbf{99.7}  & \textbf{92.0}/\textbf{97.8}  & \textbf{91.3}/\textbf{95.5}    		&\textbf{71.2}/\textbf{84.0}  &\textbf{65.2}/\textbf{80.6}  &\textbf{60.2}/75.8           			& 92.0/94.7  & 74.0/\textbf{79.3} & 71.2/\textbf{74.8}		\\



 \hline 

\end{tabular}
\begin{tablenotes}
        \footnotesize
        \item 2nd-S: Second-Stage. Mod.: Moderate. The best results are highlighted in \textbf{bold}.
\end{tablenotes}

\vspace{1mm}

\label{table:kt_abb2}
\end{table*}

\subsection{Experimental Setup}

\textbf{Datasets description}: Two widely used 3D object detection datasets (the KITTI dataset~\cite{kitti} and Waymo Open Dataset~\cite{waymo}) are chosen as the experimental database. 

The KITTI dataset consists of 7481 training frames with both LiDAR data and 3D object annotations, and 7518 test frames with only LiDAR data. Following previous work~\cite{pointpaint,pvrcnn}, the training frames are further divided into \textit{train} set with 3712 frames and \textit{val} set with 3769 frames. The objects are categorized into easy, moderate, and hard according to the box height and the occlusion ratio in the image view. 3D/BEV mean Average Precision (mAP) calculated with 11 or 40 recall positions is used as the official performance evaluation metric.

The Waymo Open Dataset includes a total of annotated 1000 sequences (around 200K frames), in which 798 sequences are taken as \textit{train} set and 202 sequences are used as \textit{val} set. 3D average precision (AP) and average precision weighted by heading accuracy (APH) are used as evaluation metrics. The dataset evaluates the results in two difficulty levels: LEVEL\_1 (L1) for object cuboids with more than 5 LiDAR points, and LEVEL\_2 (L2) for object cuboids with 1-5 LiDAR points.    

\textbf{Implementation details}: The experiments on the KITTI dataset are performed on a server with 6 NVIDIA Titan RTX GPUs, and the experiments on the Waymo Open Dataset are conducted on another server with one A800 GPU. Several advanced anchor-based LiDAR 3D object methods, including both one-stage detectors ( (SECOND~\cite{second}, PointPillars~\cite{pointpillar}, PillarNet~\cite{pillarnet}, PV-RCNN~\cite{pvrcnn}) and two-stage detectors (Focals-Conv~\cite{FSC}, Graph-Vo~\cite{graph-rcnn}, PV-RCNN++~\cite{pvrcnnplus}), are chosen as baselines. These models are re-implemented using OpenPCdet\footnote{OpenPCDet \url{https://github.com/open-mmlab/OpenPCDet}} or their original codebases, and further equipped with our proposed PASS. For a fair comparison, these models' original training settings, parameters, and inference steps (as described in their paper or code) are kept unchanged. The only difference between our models and the baselines is the adoption of the proposed PASS. Additionally, anchor-based PV-RCNN++ with PASS (named PASS-PV-RCNN++) is selected and trained as our proposed new detector, being compared with more state-of-the-art 3D object detectors~\cite{3dssd,hotspot,iassd,pointrcnn,svga-net,tanet,part2,PDV,pg-rcnn,centerpoint,gdmae,lidarrcnn} on benchmarks. In the experiments of verification of the proposed method and comparison with other state-of-the-art, $\mathcal{K}$ of PASS is set to 5. In the extended experiments, ablation studies on the effect of $\mathcal{K}$ and the number of stages are carried out.

\subsection{Verification}
To verify the effectiveness of PASS on anchor-based LiDAR 3D object detection, PASS, the comparison experiments of the models with and without PASS are conducted. Table~\ref{table:kt_val_3d} shows the experimental results on the KITTI \textit{val} set. The results are reported by the BEV mAP with 11 recall points. IoU thresholds of true positive predictions are set to 0.7 for car detection and 0.5 for pedestrian and cyclist detection. From this table, it is found that PASS can at most elevate the baseline with the mAP improvement of 1.38\% on car detection, 3.26\% on pedestrian detection, and 3.65\% on cyclist detection. Although on some metrics, PASS brings small precision drops, it achieves an average mAP improvement of at least 0.5\% and at most 1.38\% for all baseline detectors on the moderate level. Table~\ref{table:waymo_val_3d} shows the experimental results on the Waymo Open Dataset \textit{val} set. Following previous work~\cite{pvrcnn} for fast verification, only 20\% training data of \textit{train} set is used to train both baseline detectors and PASS-based detectors. From the results, PASS guarantees the mean APH (mAPH) improvement of at least 0.3\% and at most 2.0\% on all baseline models. Especially on pedestrian and cyclist detection, detectors with PASS show great performance improvement. For example, PointPillar with PASS achieves AP/APH improvement of around 3.5-4.5\% on cyclist detection. PV-RCNN++ with PASS achieves AP/APH improvement of around 0.6-1.9\% on pedestrian detection. The evaluation results prove that our proposed method can leverage the average performance of anchor-based LiDAR 3D object detection.

\subsection{Comparison with State-of-the-Art}
By applying PASS to the anchor-based version of PV-RCNN++, PASS-PV-RCNN++ is proposed. First, the total training data of the KITTI dataset is re-separated into 80\% training data and 20\% validation data. PASS-PV-RCNN++ is optimized on train data. The model with the best performance on validation data is chosen as the final detector. We evaluate the detector on the KITTI test data with the official online benchmark. It is noted that the official KITTI leaderboard is ranked by performance on moderate mAP. Since there is no anchor-based version of PV-RCNN++ on the KITTI test benchmark, we also train a PV-RCNN++ detector using the same training settings and submit it to the online benchmark. The performance of both our proposed detector and other state-of-the-art detectors submitted to online benchmark with three classes results are shown in Table~\ref{table:kt_sota_3class}. The results are computed by 3D mAP with 40 recall positions. Our proposed detector (PASS-PV-RCNN++) achieves the best average moderate mAP and cyclist moderate mAP among all anchor-based competitors, and even surpasses those advanced anchor-free detectors in the table list. Compared to the baseline PV-RCNN++, PASS enables the baseline to gain an average moderate improvement of 0.87\%. Furthermore, PASS-PV-RCNN++ is trained on the whole training data of Waymo Open Dataset and tested on the \textit{val} set. Table ~\ref{table:waymo_sota} shows the detection performance of both PASS-PV-RCNN++, PV-RCNN++, and other state-of-the-art methods. Our proposed detector (PASS-PV-RCNN++) ranks first on mAPH of object detection, AP/APH of vehicle detection, and AP/APH of cyclist detection among all anchor-based detectors, and surpasses the baseline PV-RCNN++ on mAPH with a margin of 1.0\%/0.9\%. Additionally, it exceeds some of the anchor-free detectors (e.g. CenterPoint) on vehicle detection and achieves the comparative performance of vehicle detection compared to anchor-free version of PV-RCNN++ (PV-RCNN++$^\dagger$). These comparison experimental results demonstrate that PASS can lead anchor-based LiDAR 3D object detection method to a new state-of-the-art, and shorten the performance gap between anchor-based detectors and anchor-free detectors.

\subsection{Abalation Study}
\textbf{The effect of $\mathcal{K}$}: As $\mathcal{K}$ is the only hyperparameter in the proposed PASS, here the effect of $\mathcal{K}$ is exploited based on the baseline model PointPillars~\cite{pointpillar}. Different values of $\mathcal{K}$ in
[2, 3, 4, 5, 6, 7, 8] are used to train the detector. As defined in Eq.\ref{eq:T}, small  $\mathcal{K}$  represents a large margin of dynamic boundary, while large  $\mathcal{K}$ indicates a small margin of dynamic boundary. Table~\ref{table:kt_abb} shows the experimental results. It can be seen that the detector achieves the best performance when $\mathcal{K}$ is set to 5. Overall, $\mathcal{K}$ is quite robust to the improvement of the detector.

\textbf{The effect of the number of stages}: Following Graph-Vo~\cite{graph-rcnn}, a two-stage anchor-based 3D object detector is separated into detectors without the second stage (SECOND~\cite{second}) and with the second stage (overall Graph-Vo). The effect of PASS on the different stages of the anchor-based 3D object detector is ablated in Table~\ref{table:kt_abb2}. It is observed that PASS can continuously boost the performance of the detector from only the first stage to the first stage plus the second stage, which shows that the number of stages of the 3D object detector will not restrict the application of PASS.

\subsection{Qualitative Analysis}
\label{qualitative-analysis}

\begin{figure}[t!]
\begin{center}
\includegraphics[width = 1.0\columnwidth]{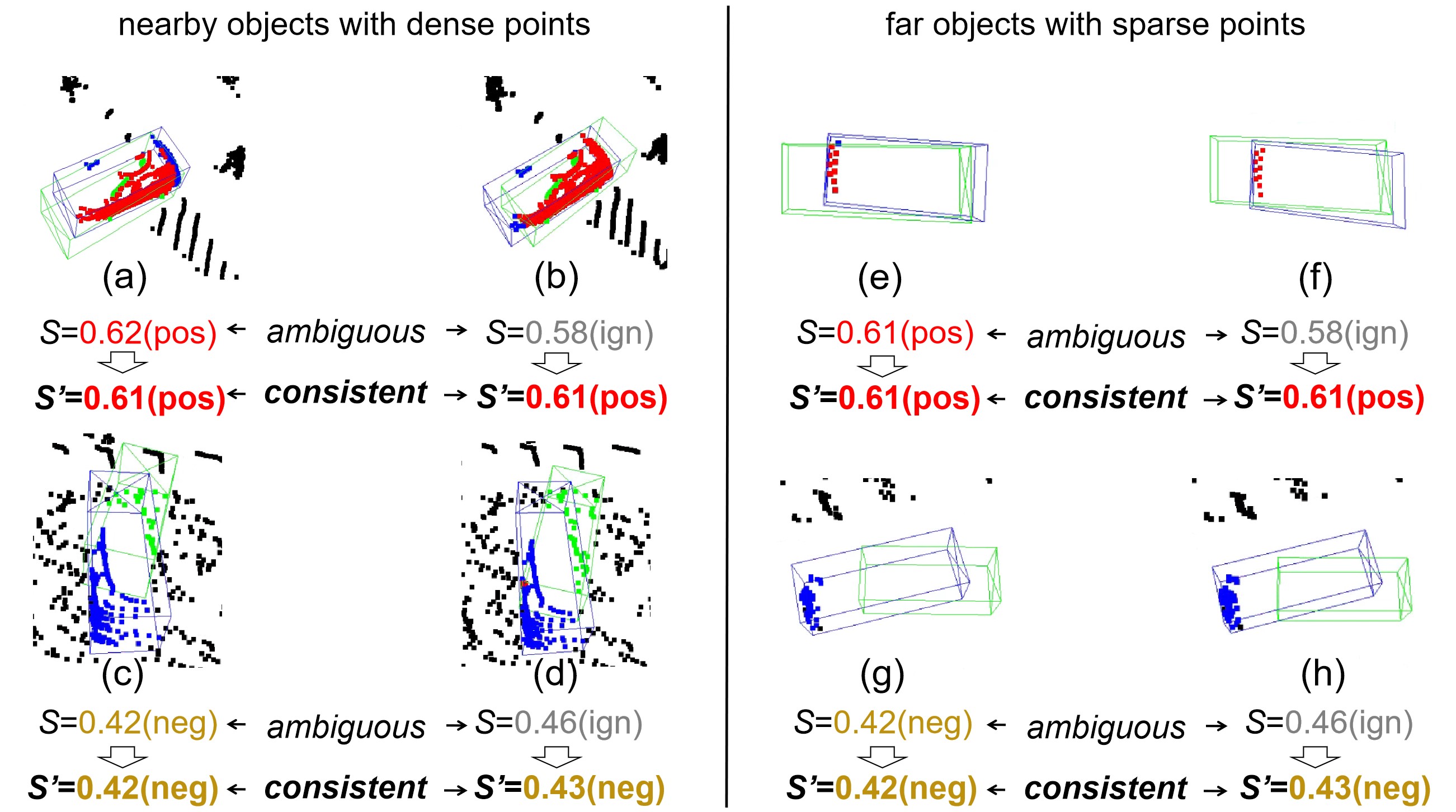}
\end{center}
    \caption{Qualitative results on training sample selection (best viewed in color). $\mathcal{S}$ and $\mathcal{S}^{\prime}$ are calculated by the IoU$_{box}$-based and the PASS-based sample selection measurement, respectively. The \textcolor{green}{green boxes} and \textcolor{blue}{blue boxes} represent the anchor samples and ground truths respectively. Points in the intersection of the anchor sample and ground truth are indicated as \textcolor{red}{red points}. Points that belong to ground truth but do not belong to the intersection are indicated as \textcolor{blue}{blue points}. Points that belong to the anchor but do not belong to the intersection are represented as \textcolor{green}{green points}, and the rest of the background points are represented as \textbf{black points}.}
\label{fig:vis_pass}
\end{figure}

\begin{figure*}[t!]
\begin{center}
\includegraphics[width = 1.0\columnwidth]{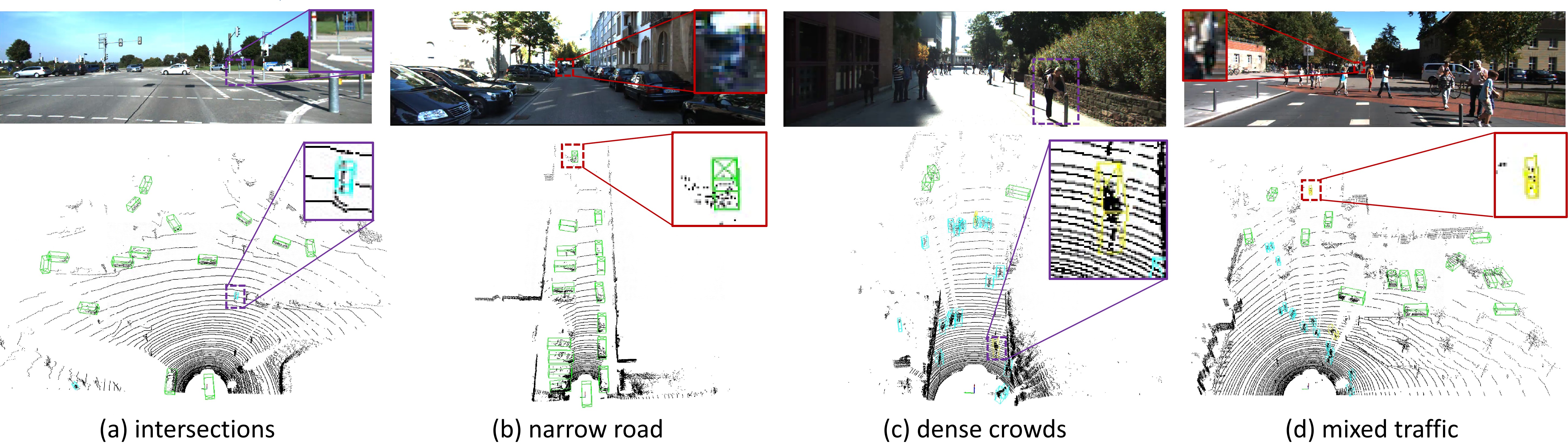}
\end{center}
    \caption{Qualitative results on challenging scenes from the KITTI \textit{test} set (best viewed in color). The top row displays the reference images, and the bottom row shows the corresponding LiDAR pointcloud with detection results from the proposed PASS-PV-RCNN++. The \textcolor{green}{green}, \textcolor{blue}{blue}, and \textcolor{yellow}{yellow} boxes represent the prediction of car, pedestrian, and cyclist respectively. Certain cases that are worth further discussing (see Section \ref{qualitative-analysis}) are highlighted by zoomed-in. }
\label{fig:vis_det}
\end{figure*}

To discover the effect of the proposed sample selection method, case studies on training sample selection are conducted on KITTI \textit{train} set. Some typical cases are visualized in figure \ref{fig:vis_pass}, including both nearby objects with dense points and far objects with sparse points. For case (a) and case (b), two anchors all contain a large number of 3D points on the same object. Based on the IoU$_{box}$-based sample selection, the anchor sample in case (a) will be divided into the positive set, while the anchor sample in case (b) will be divided into the ignored set. However, PASS will keep both two in the positive set for consistent learning. For case (c) and case (d), two anchors contain fewer object point features. The IoU$_{box}$-based sample selection will divide them into the negative set and the ignored set separately. In contrast, PASS consistently classified two anchor samples into the negative set, eliminating the ambiguity of anchor assignment. The cases (e)(f) and (g)(h) are similar to the previous examples. These cases show that PASS can better assign training samples and bring unambiguous optimization objectives to the learning of 3D object detectors.

Furthermore, the performance of the proposed PASS-based 3D object detector in complex traffic scenes is tested and visualized. Figure \ref{fig:vis_det} presents the qualitative results of the PASS-PV-RCNNN++ on the KITTI \textit{test} set. The detected bounding boxes of car, pedestrian, and cyclist are colored in green, blue, and yellow, respectively. The first row shows the referenced RGB images and the second row shows the pointclouds and 3D detection results. As observed, our PASS-based detector can make high-quality predictions in multiple challenging scenes, including intersections, narrow road, dense crowds, and mixed traffic. In figure \ref{fig:vis_det}(a), the proposed detector captures cars with different poses at intersections. However, the detector misidentified the pole in the pointcloud as a pedestrian, which reflects the difficulty of the scene, and also reflects the problem of object similarity caused by sparse points. In figure \ref{fig:vis_det}(b), the proposed detector precisely detects the occluded vehicles, conducive to the safe passage of the ego vehicle through such narrow roads. It is noted that a faraway car with sparse points is identified by the detector. In figure \ref{fig:vis_det}(c), crowded pedestrians in the scene are grabbed by the proposed detector. However, for a pedestrian with an unusual posture, the detector mistakenly recognized it as a cyclist, which mirrors the difficulty brought by complicated points deformation to 3D object detection. In figure \ref{fig:vis_det}(d), the proposed detector can detect multi-classes objects in the mixed traffic environment. It provides complete perception results for the hybrid traversal of the ego vehicle. Especially for some faraway small objects (e.g. the far cyclist in the scene), accurate category identification and 3D location positioning are all achieved by the PASS-based detector.


\subsection{Extended Experiment on PointPillars}

PointPillars~\cite{pointpillar}, as a widely-deployed anchor-based LiDAR 3D object detector, has garnered significant attention in both research and industrial domains. Numerous methods~\cite{lidarrcnn,ssl,harmonicloss} have endeavored to improve PointPillars and experimented on the KITTI \textit{test} set. In our study, we apply PASS to PointPillars and submit the detection results of the PASS-based PointPillars (PASS-PointPillars) to the KITTI \textit{test} benchmark. Table~\ref{table:kt_sota_pp} shows the car detection performance comparison of various versions of PointPillars detector. Remarkably, the proposed PASS-PointPillars achieves the best on moderate 3D mAP and easy BEV mAP among all competitors, and surpasses the baseline model with a large margin (e.g. 2.14\% improvement on easy 3D mAP). It is noteworthy that, the proposed PASS only works on the optimization of the PointPillars and introduces no extra time-cost in its inference. Specifically, it maintains a detection speed of 75.4Hz and 43.1Hz when TensorRT-based deployed and tested on Jetson Xavier TX and PC with a single 2080Ti GPU, respectively.


\subsection{Discussion}

\textbf{Manual hyperparameters}: Although PASS offers a new perspective of anchor sample selection in anchor-based LiDAR 3D object detection, it still relies on multiple manual hyperparameters (e.g. the location and shape of anchors $\left[ x^a,y^a,z^a,l^a,w^a,h^a,\theta ^a \right]$, the selection thresholds \{$\mathcal{T}_{pos}^{}$,$\mathcal{T}_{neg}^{}$\}, and $\mathcal{K}$ in PASS, etc.). These hyperparameters make the model learning inflexible. Future research could delve into these manual hyperparameters, attempting to achieve more efficient model learning with dynamically self-adaptive parameters or hyperparameter-free schemes.

\textbf{Learning efficiency}: PASS improves the performance of current anchor-based LiDAR 3D object detectors without extra inference burden. However, it does result in a training time cost increase of 1.5 to 3 times, which is unfriendly to these models with a large number of network parameters. Hence, it is worth exploring a more concise and efficient training sample selection approach in future research.

\textbf{Characteristics of LiDAR pointcloud}: Currently, many network and algorithm designs for LiDAR-based 3D vision tasks draw inspiration from or are directly adopted from camera-based 2D/3D vision methodologies. Although these designs prove effective in the image domain, they may not be inherently suitable for LiDAR pointcloud due to the different characteristics between images and pointcloud. It is worth thinking about how to design tailored methods or frameworks for 3D vision tasks based on LiDAR pointcloud.

\begin{table}[]
\caption{Performance comparison of different forms of PointPillars detector on KITTI \textit{test} benchmark. 
}
\small
\center
\renewcommand{\arraystretch}{0.85}  
\setlength\tabcolsep{2.0pt}  
\begin{tabular}{| c || c | c | c ||c | c | c |}
\hline
\multirow{2}{*}{Method}      & \multicolumn{3}{|c||}{3D} & \multicolumn{3}{|c|}{BEV}   \\ \cline{2-7}
   & Easy   & Mod.   & Hard   & Easy      & Mod.   & Hard       \\ \hline \hline


PointPillars$^\star$~\cite{pointpillar}     & 83.12	&  74.11 	&69.14	& -   & -  & -   		   
\\
PointPillars-RCNN$^\star$~\cite{lidarrcnn}        & \textbf{85.97}	&   74.21 	&69.18	& -   & -  & -   	     
\\ \hline

PointPillars$^\star$$^\star$~\cite{pointpillar}    &80.51 &68.57 &61.79 &90.74 &84.98 &79.63  	   
\\
SSL-PointPillars$^\star$$^\star$~\cite{ssl}    & 82.54	&   72.99 	&67.54	& 88.92   & 85.73  & 80.33    
\\ \hline

PointPillars$^\dagger$~\cite{pointpillar}      & 82.58	&   74.31 	&68.99	& 90.07   & 86.56  & \textbf{82.81}   	     
\\
H-PointPillars$^\dagger$~\cite{harmonicloss}        & 82.26	&   73.96 	&\textbf{69.21}	& 90.89   & \textbf{87.28}  & 82.54  		     
\\


\textbf{{PASS-PointPillars}}$^\dagger$   		 	& {84.72}  & \textbf{74.85}  & 69.05     & \textbf{91.07}  & 87.23     	& 81.98			\\

\hline


\end{tabular}
\begin{tablenotes}
        \footnotesize
        \item Mod.: Moderate. The best results are highlighted in \textbf{bold}. $\star$: results from \cite{lidarrcnn}. $\star$$\star$: results from \cite{ssl}. $\dagger$: results from KITTI test server~\cite{kitti}.
\end{tablenotes}

\vspace{1mm}

\label{table:kt_sota_pp}
\end{table}

\section{Conclusion}\label{sec5}
In this work, the ambiguity of IoU$_{box}$-based sample selection in existing anchor-based LiDAR 3D object detection methods is pointed out and thoroughly analyzed. To address this issue, a novel anchor sample selection method called PASS is proposed. PASS can be applied to any anchor-based LiDAR 3D object detector as a plug-and-play optimization part without introducing extra inference time-cost. The experimental results on multiple widely-used datasets consistently demonstrate that, with the application of PASS, anchor-based LiDAR 3D object detectors gain a great average mAP improvement, and achieve new state-of-the-art performance.


\section*{Acknowledgments}
\justifying 
This work was supported by the National Key R\&D Program of China (Grant
No. 2022YFB2502900) and the National Natural Science Foundation of China (Grant No. 62088102).
\bibliographystyle{IEEEtran}
\bibliography{kgn}
\vspace{-0.61in}

\end{document}